\documentclass[journal]{IEEEtran}
\usepackage{amssymb}
\usepackage{algorithm}
\usepackage{times}
\usepackage{url}
\usepackage[hidelinks]{hyperref}
\usepackage[utf8]{inputenc}
\usepackage[small]{caption}
\usepackage{graphicx}
\usepackage{amsmath}
\usepackage{amsthm}
\usepackage{booktabs}
\usepackage{mathrsfs}
\usepackage{algorithmic}
\floatstyle{ruled}
\restylefloat{algorithm}
\usepackage[switch]{lineno}
\usepackage{amsfonts}
\usepackage[table,xcdraw]{xcolor}
\usepackage{multirow} 
\usepackage{stfloats}
\usepackage{placeins}
\usepackage{enumitem}

\begin{document}
%
\title{Multimodal Foundation Models Adaptation based on Domain-Aware Relaxed Orthogonal Subspace for Remote Sensing}






%
%
%

\author{Han~Luo,~\IEEEmembership{Member,~IEEE,}
        Ruoyu~Yang,~\IEEEmembership{Member,~IEEE,}
        Yinhe~Liu,~\IEEEmembership{Member,~IEEE,}
        Yanfei~Zhong,~\IEEEmembership{Member,~IEEE,}

        \thanks{This work was supported in part by the National Natural Science Foundation of China under Grant 42325105 and 42501475.}
\thanks{Han Luo, Ruoyu Yang, Yinhe Liu and Yanfei Zhong are with the State Key Laboratory of Information Engineering in Surveying, Mapping and Remote Sensing, Wuhan University, Wuhan 430072, China (e-mail: luo\_han@whu.edu.cn; yangruoyu@whu.edu.cn; liuyinhe@whu.edu.cn; zhongyanfei@whu.edu.cn).}
    }
\maketitle
\begin{abstract}

Pretrained foundation models (FMs) have achieved remarkable success in computer vision, yet their high fine-tuning cost limits practical deployment. Parameter-efficient fine-tuning (PEFT) methods such as Low-Rank Adaptation (LoRA) improve efficiency by constraining updates to a predefined low-rank subspace. However, when applied to remote sensing tasks with substantial domain shifts, the fixed subspace is constructed without observing the downstream activation distribution and can therefore provide a poor coordinate system for adaptation, a phenomenon herein termed subspace mismatch. To address this issue, a unified framework is introduced, termed Domain-aware Relaxed Orthogonal Subspace adaptation (DROS), which reformulates low-rank adaptation as data-conditioned subspace learning and flexible subspace adaptation. Specifically, the weight decomposition is conditioned on second-order activation statistics estimated from the downstream training distribution, so that the initialization reflects the feature geometry actually induced by the remote-sensing data, followed by flexible geometric transformations enabled by a relaxed orthogonal parameterization. Furthermore, the framework is extended to multimodal settings (MM-DROS) by sharing transformation structures across modality-specific subspaces, facilitating efficient cross-modal interaction. Extensive experiments on multiple remote sensing benchmarks demonstrate that DROS achieves state-of-the-art performance, even surpassing full fine-tuning, without additional inference overhead.

\end{abstract}

%
\IEEEpeerreviewmaketitle
\section{Introduction}
%
%
%
%
\IEEEPARstart{F}{oundation} Models (FMs) have demonstrated general representational capabilities in the field of artificial intelligence\cite {opportunities,scale}. Vision Foundation Models (VFMs) such as CLIP\cite{clip}, DINOv2\cite{dinov2}, and EVA02\cite{fang2024eva} have achieved outstanding performance across a range of downstream tasks\cite{classification,segmentation,detection}. Concurrently, by integrating large-scale remote sensing data, Remote Sensing Foundation Models (RSFMs)\cite{gfm} have also achieved breakthroughs in remote sensing tasks with domain-specific prior knowledge such as ecology\cite{ecology}, urban planning\cite{loveda}, and earthquake recording\cite{earthquake}. Furthermore, combining multimodal satellite data with heterogeneous spatial, spectral, and temporal resolutions has proven critical, such as disaster damage assessment\cite{bright} and all-weather land-cover classification using ultra-high-resolution optical and SAR imagery\cite{whu-opt-sar}.

Lately, with the rapid development of model sizes, traditional full-parameter fine-tuning becomes impractical in real-world applications due to the immense computational and storage overhead\cite{nmi}. To address this challenge, parameter-efficient fine-tuning (PEFT) methods have emerged. In contrast to full fine-tuning, these methods significantly reduce the number of trainable parameters during fine-tuning by optimizing only the introduced lightweight adapters\cite{earth-adapter} or prompt\cite{classification}, while freezing the pretrained weights.
\begin{figure}[!tbp]
    \centering
    \includegraphics[width=\columnwidth]{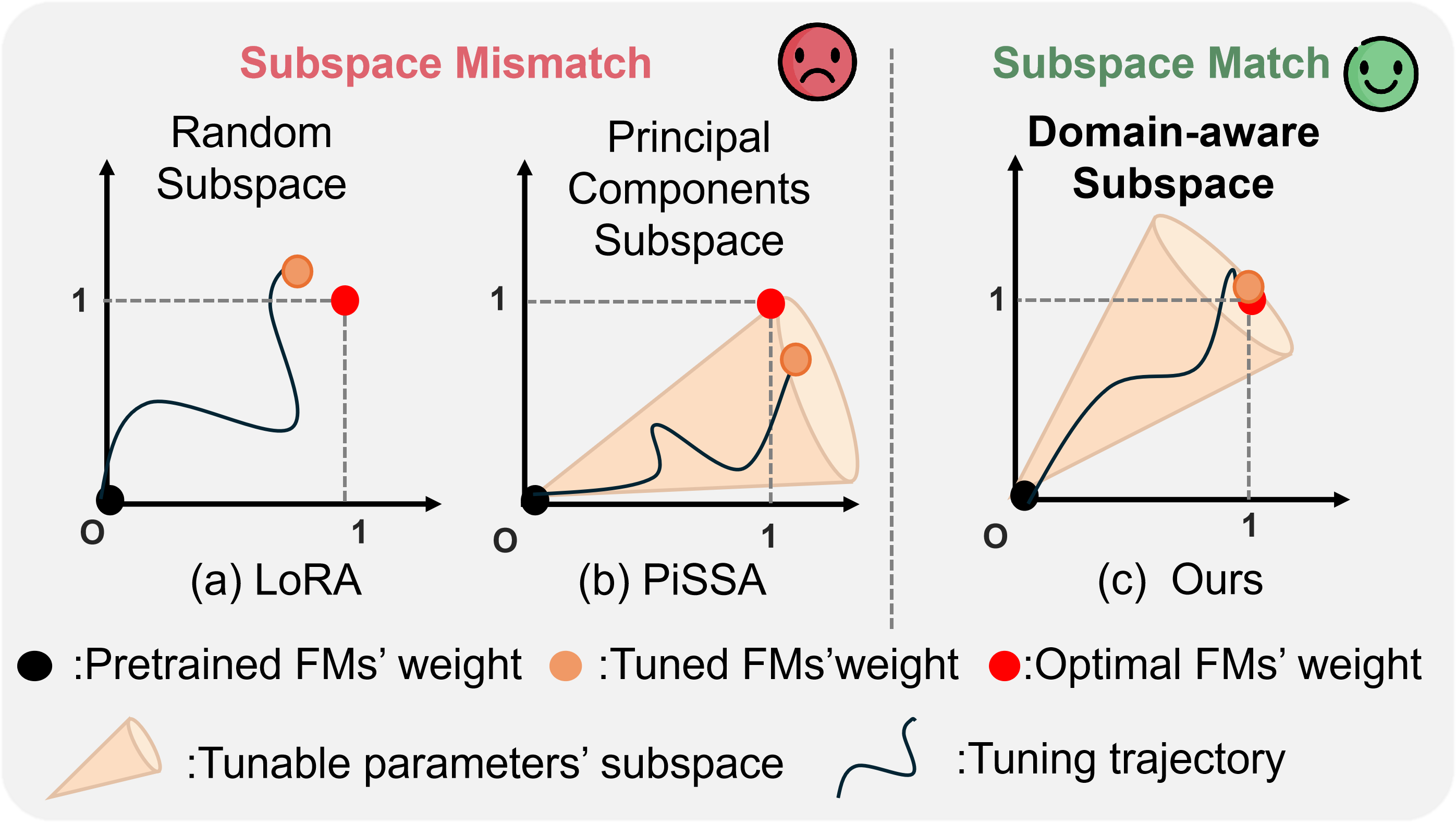}
    \caption{Conceptual illustration of \textit{subspace mismatch} in parameter-efficient fine-tuning (PEFT) under domain shifts. (a) LoRA\cite{lora} constructs a low-rank update within a randomly initialized subspace without using downstream activation statistics. (b) PiSSA\cite{pissa} derives the adaptation subspace from pretrained weights, improving optimization efficiency but remaining independent of the downstream activation distribution. (c) DROS conditions the subspace initialization on second-order activation statistics from representative training data and further enables structured transformations within this data-conditioned subspace.}
    \label{fig:radar}
\end{figure}

Particularly, low-rank adaptation (LoRA)\cite{lora} suggests that weight updates during FMs' fine-tuning exhibit a low-rank structure, and accordingly employs two low-rank matrices with low intermediate dimensions as learnable adapters for each linear layer. From a linear algebra perspective, this constrains fine-tuning to find optimal solutions within a predefined low-rank subspace. Up to now, LoRA has gained increasing popularity as its low-rank matrices can be incorporated into the original linear weights after fine-tuning, ensuring that the model architecture remains unchanged and incurs no additional inference overhead\cite{dora}. Subsequent research has enhanced LoRA by using adaptive intermediate low ranks across layers to dynamically adjust the subspace\cite{adalora}, or by defining different subspaces\cite{pissa, dora, oftv2} and reducing the number of trainable parameters within predefined subspaces further\cite{lora-xs}. 

As illustrated in Fig.~\ref{fig:radar}, existing low-rank PEFT methods differ mainly in how the adaptation subspace is initialized. LoRA\cite{lora} starts from randomly initialized low-rank factors, whereas PiSSA\cite{pissa} derives the initial coordinates from the principal components of the pretrained weights. Although these strategies introduce different initialization priors, they share a common characteristic: the initial adaptation coordinates are determined without considering the feature responses produced by the data actually used for downstream adaptation. Because the adaptation rank is usually small, the initialization determines how the limited low-rank capacity is allocated. A data-independent initialization may therefore assign this capacity to directions that are not sufficiently informative for the observed downstream data.

This limitation is particularly relevant to remote sensing. Even when the semantic task remains unchanged, variations in season and phenology, sensor characteristics, and geographical scene composition can substantially change the observed imagery. These variations are further propagated through the frozen foundation model, leading to different downstream activation distributions. Consequently, an adaptation subspace determined by a predefined initialization may not adequately reflect the feature distribution encountered during fine-tuning. This disconnect between data-agnostic subspace initialization and the observed downstream activation distribution is referred to as \textit{subspace mismatch}.

CorDA\cite{corda} provides a relevant methodological precedent for addressing subspace mismatch. In large language models, CorDA introduced context-oriented decomposition by estimating input activation covariance from representative samples and incorporating it into the decomposition of pretrained linear weights. This finding suggests that the data observed during adaptation can provide useful guidance for determining where the limited low-rank adaptation capacity should be allocated. However, whether this context-conditioned decomposition principle can be effectively transferred to remote-sensing foundation models remains insufficiently explored, particularly because remote sensing imagery is affected by heterogeneous environmental and acquisition conditions.

However, identifying data-relevant adaptation directions is only the first step; under a small rank, the limited adaptation capacity must also be effectively utilized within the selected subspace. Although covariance-conditioned decomposition provides a data-driven initialization of the low-rank bases, the conventional formulation $\Delta W=BA$ provides limited flexibility for combining and transforming the selected directions during adaptation\cite{mtlora}. MTLoRA\cite{mtlora} showed that inserting an intermediate transformation matrix, i.e., $\Delta W=BTA$, enables more flexible geometric transformations within the low-rank space. This observation motivates the introduction of an explicit within-subspace operator. Nevertheless, an unconstrained transformation may excessively perturb the structure inherited from pretraining, whereas a strictly orthogonal transformation preserves geometric structure but lacks adaptive scaling. A suitable operator should therefore balance structural preservation with sufficient adaptation flexibility.

Motivated by these two complementary considerations, Domain-aware Relaxed Orthogonal Subspace adaptation (DROS) is proposed. First, DROS estimates per-layer activation covariance from a small representative subset of downstream training images and performs SVD on the covariance-conditioned weight $W^{\mathrm{pre}}C$, transferring the context-oriented decomposition principle of CorDA to remote-sensing foundation models. The bottom-$r$ components are selected to initialize adaptation while preserving the dominant pretrained components. Second, DROS introduces a relaxed orthogonal operator between the two low-rank bases, enabling structured transformation and adaptive scaling within the selected subspace. In this way, covariance-conditioned decomposition determines where the limited adaptation capacity is allocated, while the relaxed orthogonal operator improves how effectively that capacity is utilized. Here, \emph{domain-aware} means that the subspace initialization is conditioned on activation statistics estimated from the available training data; it does not imply explicit identification of the physical source of a distribution shift or direct alignment with an unseen domain. The same principle is further extended to multimodal remote sensing as MM-DROS, where optical and SAR branches retain modality-specific bases while sharing the compact transformation operator.

In summary, the main contributions are as follows:
\begin{itemize}[leftmargin=*]
\item \textbf{Subspace mismatch} is identified in remote-sensing PEFT, where data-agnostic initialization may allocate limited low-rank capacity to directions that do not reflect the observed downstream activation distribution.
\item \textbf{DROS} is proposed to allocate low-rank capacity to data-informed directions through covariance-conditioned decomposition, while a relaxed orthogonal operator improves the utilization of the selected subspace without disrupting dominant pretrained representations.
\item DROS is extended to multimodal settings as \textbf{MM-DROS}, which retains modality-specific bases while sharing the compact transformation operator across modalities.
\end{itemize}
Extensive experiments on VFMs and RSFMs demonstrate consistent improvements over representative PEFT baselines across multiple remote-sensing tasks, without additional inference latency.
\section{Related Work}
\subsection{Foundation model adaptation in remote sensing}
The concept of foundation models was introduced in natural language processing (NLP)\cite{opportunities}, defined as large-scale models pretrained on massive data and adaptable to a wide range of downstream tasks. In vision, backbones such as ViT\cite{vit}, Swin Transformer\cite{swin}, DINOv2\cite{dinov2}, CLIP\cite{clip}, SAM\cite{sam}, and EVA02\cite{eva,fang2024eva} learn transferable representations from large-scale data, and remote-sensing foundation models (RSFMs)\cite{gfm} adapt this paradigm to Earth observation. Representative RSFMs exploit contrastive learning over spatio-temporal structures, as in GASSL\cite{gassl}, SeCo\cite{seco}, and CACo\cite{caco}; masked image modeling, as in Scale-MAE\cite{scale-mae}, RingMo\cite{ringmo}, and RVSA\cite{rvsa}; model scaling\cite{billion-rs}; and multimodal pretraining, as in CROMA\cite{croma}, DeCUR\cite{decur}, AnySat\cite{anysat}, and DOFA\cite{dofa}. MARS\cite{mars} further demonstrates strong multi-source capabilities, and recent task-specific works continue to advance remote sensing image processing, e.g., SCRKD\cite{scrkd} for cross-resolution salient object detection. Despite this progress, efficiently adapting these large models to the numerous and often domain-shifted remote sensing tasks remains an open challenge: full fine-tuning is computationally prohibitive, and the representation gap between the pretraining distribution and remote sensing imagery is substantial. This motivates parameter-efficient yet domain-aware adaptation strategies.
\subsection{Low-rank PEFT and subspace initialization}
Low-rank adaptation\cite{lora} freezes pretrained weights and learns a low-rank update, and subsequent work focuses on how the adaptation subspace is initialized and structured. DoRA\cite{dora} decouples magnitude and direction, OFT\cite{oftv2} enforces orthogonality to preserve pretrained geometry, and LoRA-XS\cite{lora-xs}, AdaLoRA\cite{adalora}, and MTLoRA\cite{mtlora} refine the low-rank structure or its adaptivity. PiSSA\cite{pissa} initializes the update along the principal components of the pretrained weights, which accelerates early convergence but does not use downstream activation statistics. ASVD\cite{asvd} further shows that weight decomposition can benefit from accounting for the activation distribution. Most directly related to our first stage, CorDA\cite{corda} performs context-oriented decomposition in large language models by multiplying each pretrained linear weight by the input-activation covariance estimated from representative samples before SVD; in this way, the factorizing orientation depends on the data context. DROS transfers this covariance-conditioned decomposition principle to remote-sensing foundation models, where the relevant context is a downstream image distribution affected by acquisition and scene variation. Unlike CorDA, DROS uses the bottom-$r$ components for semantic preservation, adds a relaxed orthogonal operator for structured within-subspace transformation, and extends the formulation to multimodal remote sensing. The contribution is therefore not the claim that covariance uniquely identifies a physical domain, but that downstream activation statistics can provide a compact conditioning signal for constructing the adaptation coordinates.
\subsection{Remote sensing PEFT under domain shift}
Several works adapt PEFT to remote sensing. AiRs\cite{airs} designs spatial-context and semantic-response adapters for remote sensing visual tasks, and TEA\cite{tea} improves adaptability through residual connections while reducing training memory via gradient highways; UPetu\cite{upetu} combines quantization with learnable prompts; DEFLECT\cite{deflect} adapts RGB-pretrained models to multispectral inputs; Earth-Adapter\cite{earth-adapter} mixes frequency-domain adapter experts to suppress image artifacts; and prompt-based methods\cite{classification,vpt} inject learnable prompts into frozen backbones. The rapid adoption of parameter-efficient strategies for remote sensing foundation models has been systematically reviewed in a recent survey\cite{rseftuning}, and this line of work continues to grow, e.g., PeftCD\cite{peftcd} applies LoRA and adapters to change detection. These methods improve remote-sensing adaptation through architectural or optimization priors, but the low-rank coordinate system in the compared PEFT baselines is not explicitly conditioned on the downstream activation statistics. DROS focuses on this complementary question by using second-order activation statistics from representative training samples to condition the weight decomposition, while retaining mergeability and therefore no additional inference latency, as reported in Tab.~\ref{tab:efficiency}.
\subsection{Multimodal remote sensing adaptation}
Multimodal remote sensing, e.g., combining optical and SAR imagery, is central to applications such as disaster assessment. Multimodal pretraining methods such as CROMA\cite{croma}, DeCUR\cite{decur}, AnySat\cite{anysat}, and DOFA\cite{dofa} learn joint representations, while adapter-based methods fine-tune models such as SAM\cite{sam} for specific multimodal tasks, including RSPrompter\cite{rsprompter}, SAM-RSIS\cite{sam-rsis} built on the ViT adapter\cite{vit-adapter}, and Classwise-SAM-adapter\cite{classwise}. For multimodal remote sensing object detection, DPAL\cite{dpal} further tackles cross-modal and task-head misalignment through dual-perspective alignment learning. However, these methods typically adapt each modality independently, ignoring the pronounced heterogeneity between optical and SAR imagery---different measurement principles, intensity statistics, and noise patterns---which leads to poorly aligned per-modality representations. MM-DROS addresses this by sharing a relaxed orthogonal operator across modality-specific subspaces, coupling the transformations of the two modalities and enabling efficient cross-modal gradient interaction.
\begin{figure*}[!tbp] 
    \centering
    \includegraphics[width=\textwidth]{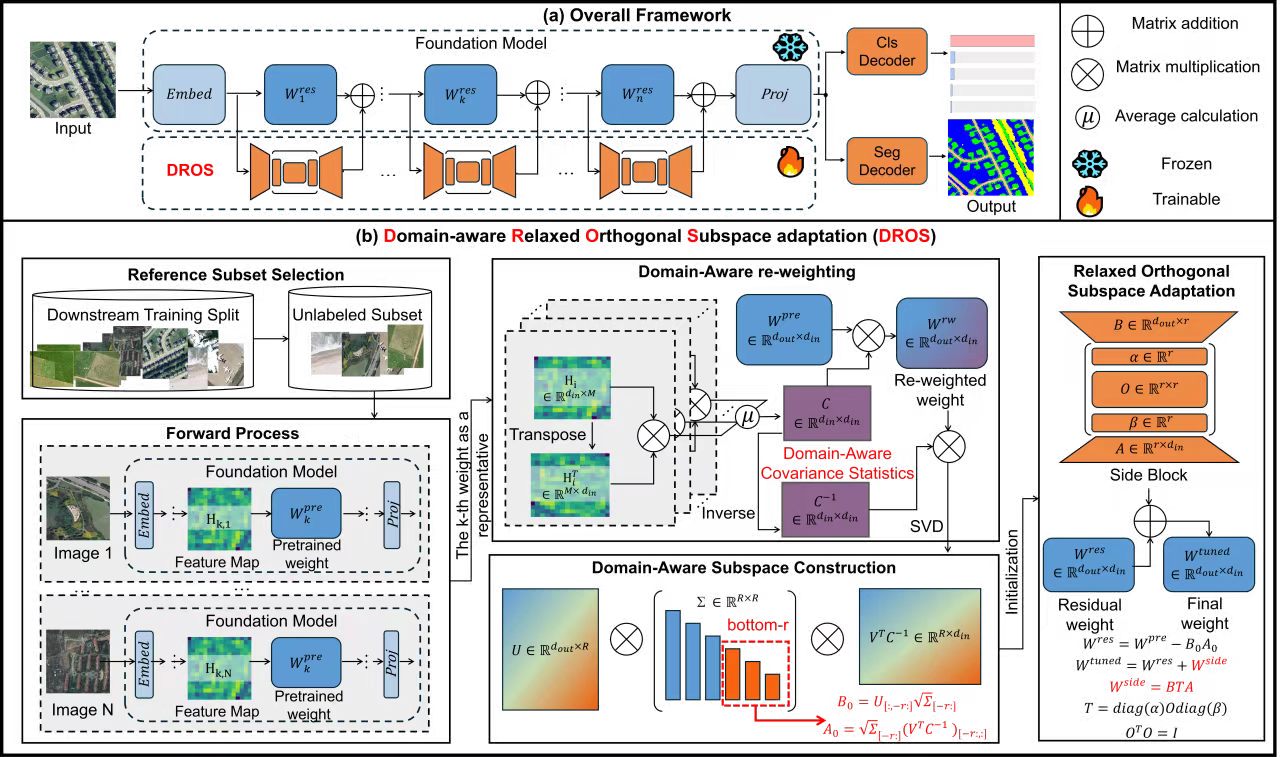}
    \caption{Overview of DROS. Panel a presents the overall framework, where a trainable DROS side branch is attached to each of the $n$ adapted linear layers in a frozen foundation model. Panel b details the construction for layer $k$, where $k\in\{1,\ldots,n\}$. The frozen backbone collects $H_{k,i}$ from each of $N$ representative training samples, and averaging the products $H_{k,i}H_{k,i}^{\top}$ gives $C_k$. The index $k$ is omitted after a single layer is fixed. Covariance-conditioned SVD of $W^{\mathrm{rw}}=W^{\mathrm{pre}}C$ selects the bottom-$r$ components to initialize $B_0$ and $A_0$. The pretrained weight is split as $W^{\mathrm{pre}}=W^{\mathrm{res}}+B_0A_0$, and the trainable side block $W^{\mathrm{side}}=B\mathcal{T}A$ yields $W^{\mathrm{tuned}}=W^{\mathrm{res}}+W^{\mathrm{side}}$.}
    \label{fig:large_architecture}
\end{figure*}
\section{Method}
This section presents the DROS framework for efficient and robust foundation model adaptation in remote sensing. Sec. III-A introduces PEFT preliminaries and formalizes subspace mismatch. Sec. III-B presents the \textit{domain-aware subspace learning} module for constructing domain-aware subspaces via covariance statistics. Sec. III-C introduces the \textit{relaxed orthogonal subspace adaptation} module for enhancing geometric flexibility within the learned subspace. Sec. III-D extends the framework to multimodal settings as MM-DROS.
\subsection{Problem Formulation}
Parameter-Efficient Fine-Tuning methods aim to adapt pretrained models to new tasks by updating only a small number of parameters, rather than the entire model, thereby preventing overfitting while improving performance. Given a pretrained model $M$ parametrized by $\theta$, and a downstream task $\mathcal{D} = {(x_i, y_i)}^{\lvert \mathcal{D}\rvert}_{i=1}$, where $(x_i, y_i)$ serves as a ground truth input-output pair of task $\mathcal{D}$, parameter-efficient fine-tuning aims to adapt $\theta$ to task $\mathcal{D}$, where task-specific parameters increment $\Delta\theta$ is introduced with $\lvert\Delta\theta\rvert \ll\lvert\theta\rvert$. The optimal parameters are found by optimizing the losses $\mathcal{L}$ on task $\mathcal{D}$:
\begin{equation}
\begin{aligned}
\min_{\Delta \theta} \mathbb{E}_{(x_i, y_i) \in \mathcal{D}}\mathcal{L}\left(M_{\theta+\Delta \theta}(\hat{y}_i|x_i), y_i\right).
\end{aligned}
\label{eq:peft}
\end{equation}

LoRA observes that the weight updates after LLM fine-tuning exhibit an inherently low-rank structure\cite{lora}. Based on this insight, it approximates the weight change using two learnable low-rank matrices while keeping the pretrained weights frozen throughout the fine-tuning process. For updating pretrained weight matrix $W^{\mathrm{pre}}\in\mathbb{R}^{d_{\mathrm{out}}\times d_{\mathrm{in}}}$ to obtain $W=W^{\mathrm{pre}}+\Delta W$, the update $\Delta W$ is parameterised by a low-rank decomposition $BA$ as Eq. \ref{eq:lora}:
\begin{equation}
\left\{
\begin{aligned}
&h = Wx = (W^{\mathrm{pre}} + BA)x, \\
&\operatorname{rank}(B) = \operatorname{rank}(A) = r
\end{aligned}
\right.
\label{eq:lora}
\end{equation}
where $B \in \mathbb{R}^{d_{\mathrm{out}} \times r}$ and $A \in \mathbb{R}^{r \times d_{\mathrm{in}}}$ are trainable matrices. Following standard practice, $A$ is initialised with Kaiming initialisation and $B$ with zeros, so training begins from $W^{\mathrm{pre}}$. After fine-tuning, the product $BA$ can be incorporated into the pretrained weight W without altering the model architecture and incurring any additional inference cost.

Building upon this low-rank paradigm, PiSSA\cite{pissa} reinterprets the initialisation of $A$ and $B$ from the perspective of Singular Value Decomposition (SVD) as Eq.\ref{eq:pissa}:
\begin{equation}
\left\{
\begin{aligned}
&W^{\mathrm{pre}} = U\Sigma V^\top = W^{\mathrm{res}}+BA, \\
&A = \sqrt{\Sigma_r} V_r^\top, B = U_r \sqrt{\Sigma_r} \\
&h=(W^{\mathrm{res}}+BA)x
\end{aligned}
\right.
\label{eq:pissa}
\end{equation}
where $U_r \in \mathbb{R}^{d_{\mathrm{out}} \times r}$, $\Sigma_r \in \mathbb{R}^{r \times r}$, and $V_r^\top \in \mathbb{R}^{r \times d_{\mathrm{in}}}$ represent the top-$r$ principal singular components that need to be fine-tuned. However, it remains fundamentally domain-agnostic, as the SVD is performed on the general pretrained weights without accounting for the specific spatial pattern of remote sensing data.
\subsection*{Overall framework}
To address the subspace misalignment and limited geometric flexibility in existing PEFT methods, low-rank adaptation is reformulated as a two-stage process in DROS: domain-aware subspace learning, followed by relaxed orthogonal subspace adaptation. As illustrated in the upper path of Fig.~\ref{fig:large_architecture}, DROS is applied to the side branches of all adapted linear layers in FMs.

Let $n$ denote the number of adapted backbone linear layers, and let $k\in\{1,\ldots,n\}$ index one such layer. The statistic $C_k$ is estimated from the input activations of layer $k$ and determines its covariance-conditioned initialization factors $B_{k,0}$ and $A_{k,0}$.

The first stage focuses on data-conditioned subspace construction. Unlike LoRA and PiSSA, whose initialization does not depend on downstream activations, DROS uses the per-layer statistics $\{C_k\}_{k=1}^{n}$ derived from a representative subset of the downstream training data to condition the weight decomposition. Consequently, the orientation of the initial low-rank coordinates depends on the activation geometry observed in the downstream distribution. The term \emph{domain-aware} is used for this conditioning effect; it does not imply exact identification of the domain or guaranteed recovery of a unique optimal subspace.

The second stage addresses subspace flexibility. To overcome the geometric rigidity of fixed low-rank updates, a relaxed orthogonal transformation $\mathcal{T}$ is introduced between the learned bases. Parameterized by an orthogonally constrained matrix with relaxation factors, $\mathcal{T}$ enables steerable geometric operations within the subspace, thereby improving expressivity while preserving pretrained knowledge.

This formulation separates data-conditioned subspace initialization from the subsequent transformation performed within that subspace. For compactness, Eq.~\ref{eq:overall_dros} and the subsequent single-layer derivations fix one adapted layer and suppress the index $k$. Fig.~\ref{fig:large_architecture} shows both the layer-wise and fixed-layer conventions, while Algorithm~\ref{alg:road_standard} retains $k$ for the implementation. Let $B_0$ and $A_0$ denote the covariance-conditioned factors at initialization, and let $W^{\mathrm{res}}$ denote the corresponding frozen residual weight. The overall formulation is
\begin{equation}
\left\{
\begin{aligned}
&W^{\mathrm{pre}} = W^{\mathrm{res}}+B_0A_0, \\
&W^{\mathrm{tuned}} = W^{\mathrm{res}}+W^{\mathrm{side}}, \\
&W^{\mathrm{side}} = B\mathcal{T}A, \\
&\mathcal{T} = \operatorname{diag}(\alpha)O\operatorname{diag}(\beta).
\end{aligned}
\right.
\label{eq:overall_dros}
\end{equation}
The trainable factors $B$ and $A$ start from $B_0$ and $A_0$. Initialization also sets $O=I_r$ and $\alpha=\beta=\mathbf{1}_r$, which gives  $\mathcal{T}=I_r$.   The detailed formulations are presented in the lower path of Fig.~\ref{fig:large_architecture} and the following sections.
\subsection{Domain-aware Subspace Learning}
The covariance-conditioned construction is motivated by CorDA\cite{corda}, which introduced context-oriented decomposition for large language models. CorDA estimates input-activation covariance from representative samples and uses it to condition the SVD of pretrained linear weights, allowing the decomposition basis to depend on the observed data context. Following this principle, DROS uses representative downstream training images as the remote sensing context and conditions each adapted layer on the corresponding activation statistic. The goal is not to infer which physical factor caused a distribution shift, but to make the low-rank initialization responsive to the feature distribution actually observed during adaptation. The construction consists of estimating a per-layer second-order activation statistic, conditioning the weight decomposition on this statistic, and selecting the components used to initialize adaptation.
\subsubsection{Domain-aware statistics and weight decomposition}
The lower path of Fig.~\ref{fig:large_architecture} illustrates the construction of domain-aware covariance statistics. Let $N=\lvert\mathcal{D}_{\mathrm{sub}}\rvert$ denote the number of representative samples. For layer $k$ and sample $i\in\{1,\ldots,N\}$, the layer-input activation map is arranged as $H_{k,i}\in\mathbb{R}^{d_{\mathrm{in},k}\times M_k}$. Each of its $M_k$ columns corresponds to a token, patch, or spatial position. Under the fixed-layer convention, $H_i\equiv H_{k,i}$, $M\equiv M_k$, $d_{\mathrm{in}}\equiv d_{\mathrm{in},k}$, and $C\equiv C_k$. The layer statistic is the uncentered second-order activation statistic
\begin{equation}
\begin{aligned}
C = \frac{1}{N} \sum_{i=1}^{N} H_iH_i^\top \in \mathbb{R}^{d_{\mathrm{in}}\times d_{\mathrm{in}}}.
\end{aligned}
\label{eq:covariance}
\end{equation}
Thus, $C_k$ is defined separately for every adapted layer, whereas $C$ denotes the statistic of the fixed layer in the derivation. The symbol $C^{-1}$ denotes the inverse of that layer statistic. Following the covariance-conditioned decomposition shown in Fig.~\ref{fig:large_architecture}, the pretrained weight matrix $W^{\mathrm{pre}}$ is reweighted by $C$ and decomposed via SVD:
\begin{equation}
\begin{aligned}
W^{\mathrm{rw}} = W^{\mathrm{pre}} C = U \Sigma V^\top
\end{aligned}
\label{eq:re-weighted}
\end{equation}
where $W^{\mathrm{rw}}$ denotes the covariance-conditioned weight matrix that incorporates second-order statistics of input features. Let $R=\min\{d_{\mathrm{out}},d_{\mathrm{in}}\}$ denote the compact SVD dimension, with $r\ll R$. Thus, $U\in\mathbb{R}^{d_{\mathrm{out}}\times R}$ and $V^\top C^{-1}\in\mathbb{R}^{R\times d_{\mathrm{in}}}$. This decomposition yields a set of basis vectors ordered by their contribution under the covariance-induced transformation, which serves as the foundation for subsequent subspace selection.
\subsubsection{Minor component selection and subspace initialization}
In the SVD of the covariance-conditioned matrix $W^{\mathrm{rw}}=W^{\mathrm{pre}}C$, the components are ordered by their contribution to the mapping jointly determined by the pretrained weight and the observed activation statistics. DROS adopts a preservation-oriented selection rule: the top components, which make the largest contribution to this covariance-conditioned mapping, are retained in the frozen residual weight, while the trainable branch is initialized from the bottom-$r$ components. The purpose is not to maximize early convergence speed, but to reduce direct perturbation of the dominant pretrained structure while allocating adaptation capacity to lower-energy residual directions. Which differs from PiSSA\cite{pissa} adapting its principal components to improve early optimization. 

Specifically, the pretrained weight matrix is decomposed as:
\begin{equation}
\left\{
\begin{aligned}
& B_0 = U_{[:, -r:]} \sqrt{\Sigma_{[-r:]}}\quad
A_0 = \sqrt{\Sigma_{[-r:]}} (V^\top C^{-1})_{[-r:, :]}\\
& W^{\mathrm{res}} = W^{\mathrm{pre}} - B_0A_0
\end{aligned}
\right.
\label{eq:ba}
\end{equation}
where $C^{-1}$ maps the right singular basis of $W^{\mathrm{rw}}$ back to the original input coordinates, as depicted in Fig.~\ref{fig:large_architecture}. The trainable factors are initialized as $B\leftarrow B_0$ and $A\leftarrow A_0$, while $W^{\mathrm{res}}$ is frozen. Equation~\ref{eq:ba} therefore supplies the initialization used in Eq.~\ref{eq:overall_dros}.

During optimization, parameter updates are confined to the subspace spanned by the selected singular vectors, which provides a structured parameterization consistent with the decomposition induced by the covariance matrix.
\subsection{Relaxed orthogonal subspace adaptation}
Given the constructed subspace, adaptation is performed within this space through a structured transformation of the parameters. The formulation is based on orthogonality constraints, followed by a relaxed parameterization that allows flexible updates while preserving the underlying structure.
\subsubsection{Orthogonality-Constrained Formulation}
To introduce structured transformations within the low-rank subspace, an orthogonal matrix $O \in \mathbb{R}^{r \times r}$ is inserted between the low-rank factors:
\begin{equation}
\begin{aligned}
W^{\mathrm{tuned}} = W^{\mathrm{res}} + B O A, \quad \text{s.t. } O^\top O = I_r
\end{aligned}
\label{eq:oad}
\end{equation}
where $W^{\mathrm{tuned}}$ is the weight after fine-tuning. The matrix $O$ is parameterized via the Cayley transform:
\begin{equation}
O = (I + Q)(I - Q)^{-1}, \quad Q = -Q^\top.
\end{equation}
The matrix $O$ is not optimized directly. The trainable parameters instead define the skew-symmetric matrix $Q$, from which $O$ is reconstructed at each forward pass. An exact inverse gives $O^\top O=I_r$ and preserves inner products. The implementation evaluates the inverse term with a fifth-order truncated Neumann series following OFT-v2\cite{oftv2}, producing the intended near-orthogonal structure without explicit projection.
\subsubsection{Relaxed orthogonal parameterization}
Strict orthogonality constrains the operator $O$ to norm-preserving transformations and does not provide explicit coordinate-wise magnitude modulation. Therefore, as illustrated in the lower-right part of Fig.~\ref{fig:large_architecture}, two learnable scaling vectors $\alpha,\beta \in \mathbb{R}^r$ are introduced on the two sides of $O$:
\begin{equation}
\begin{aligned}
W^{\mathrm{tuned}} = W^{\mathrm{res}} + B \left( \operatorname{diag}(\alpha)\, O \, \operatorname{diag}(\beta) \right) A
\end{aligned}
\label{eq:road}
\end{equation}
Let $D_{\alpha}=\operatorname{diag}(\alpha)$ and $D_{\beta}=\operatorname{diag}(\beta)$. In $D_{\alpha}OD_{\beta}$, $\beta$ modulates the latent coordinates before orthogonal reorientation, whereas $\alpha$ modulates the transformed coordinates afterward. For a general non-diagonal $O$, diagonal scaling does not commute with the orthogonal transformation; therefore, pre- and post-reorientation scaling are not equivalent at the operator level and, in general, cannot be replaced by a single diagonal scaling while keeping $O$ fixed. Nevertheless, because the surrounding low-rank factors $A$ and $B$ are also trainable, the complete mapping contains scale-equivalent parameterizations, e.g., $BD_{\alpha}OD_{\beta}A=(BD_{\alpha})O(D_{\beta}A)$. Accordingly, the dual-sided form is not intended to enlarge the theoretical function class, but is used as a structured optimization parameterization that explicitly separates magnitude modulation before and after orthogonal coordinate mixing. To sum up, the training starts from $B=B_0$, $A=A_0$, $Q=0$, and $\alpha=\beta=\mathbf{1}_r$, which give $O=I_r$ and $\mathcal{T}=I_r$, satisfying the pretrained recovery established in Eq.~\ref{eq:overall_dros}. The entire pipeline is described by Algorithm~\ref{alg:road_standard}.
\begin{figure}[!tbp] 
    \centering
    \includegraphics[width=0.75\columnwidth]{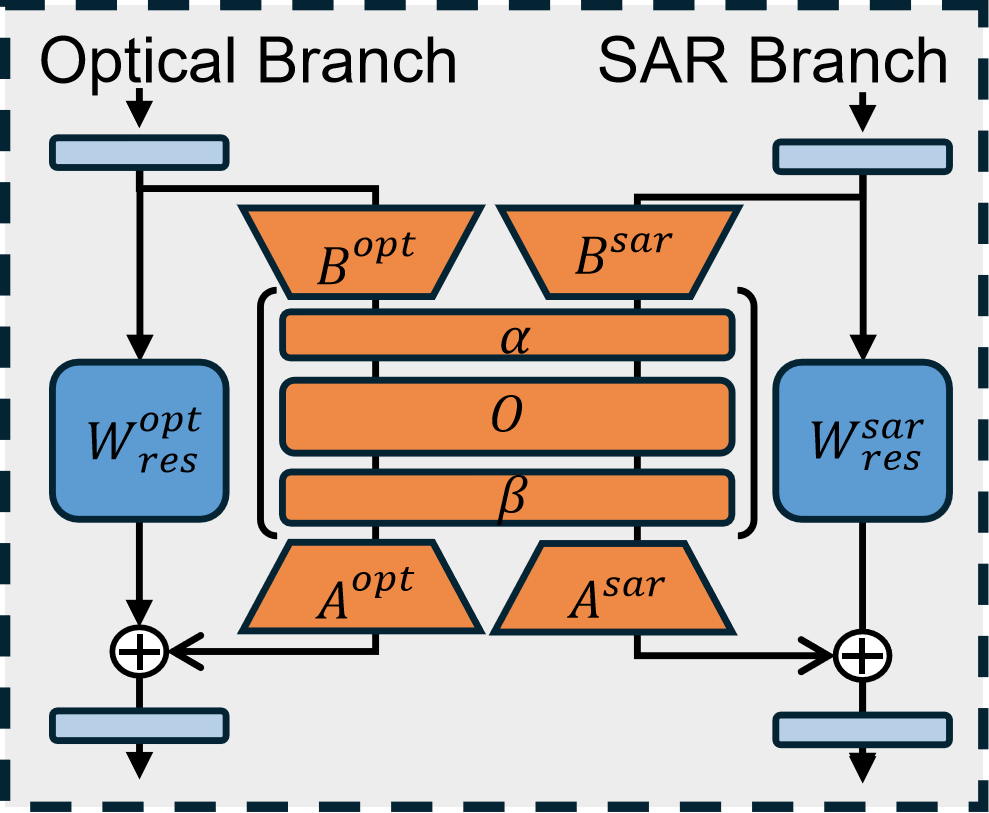}
    \caption{Overview of MM-DROS at one adapted linear layer. The layer index $k$ is suppressed. The optical and SAR branches retain modality-specific residual weights and bases $B^{\mathrm{opt}},A^{\mathrm{opt}}$ and $B^{\mathrm{sar}},A^{\mathrm{sar}}$. The relaxed operator $\mathcal{T}_{\mathrm{shared}}=\operatorname{diag}(\alpha)O\operatorname{diag}(\beta)$ is shared across the two branches. Each transformed low-rank branch is added to its corresponding residual weight.}
	\label{fig:mmroad}
\end{figure}
\subsection{Extension to multimodal domain-aware adaptation}
To address cross-modal discrepancies between optical and SAR modalities, the proposed framework is extended to a multimodal setting, termed MM-DROS. As illustrated in Fig.~\ref{fig:mmroad}, two modality-specific branches are constructed, each equipped with its own domain-aware subspace bases, i.e., $\{B^{opt}, A^{opt}\}$ and $\{B^{sar}, A^{sar}\}$. These bases are independently derived to capture modality-dependent characteristics. Within each branch, the relaxed orthogonal parameterization in Eq.\ref{eq:road} is adopted. To enable interaction across modalities, the transformation operator is shared between the two branches. Specifically, a shared operator $\mathcal{T}_{shared}$ is imposed at the center of both DROS structures, while the surrounding subspace bases remain modality-specific. The adaptation process is formulated as:
\begin{equation}
\begin{cases}
W_{\mathrm{sar}}^{\mathrm{tuned}} = W_{\mathrm{sar}}^{\mathrm{res}} + B^{\mathrm{sar}}\mathcal{T}_{\mathrm{shared}}A^{\mathrm{sar}} \\
W_{\mathrm{opt}}^{\mathrm{tuned}} = W_{\mathrm{opt}}^{\mathrm{res}} + B^{\mathrm{opt}}\mathcal{T}_{\mathrm{shared}}A^{\mathrm{opt}}
\end{cases}
\label{eq:mmroad}
\end{equation}
In Eq.~\ref{eq:mmroad}, the indices $\mathrm{opt}$ and $\mathrm{sar}$ denote modality-specific branches. Consistent with the fixed-layer convention, the index $k$ is omitted. For each modality $m\in\{\mathrm{opt},\mathrm{sar}\}$, the frozen residual weight is $W_m^{\mathrm{res}}=W_m^{\mathrm{pre}}-B_0^mA_0^m$. The operator $\mathcal{T}_{\mathrm{shared}}=\operatorname{diag}(\alpha)O\operatorname{diag}(\beta)$ is shared across modalities, while $B^m$ and $A^m$ remain modality-specific and trainable. Initialization sets $B^m=B_0^m$ and $A^m=A_0^m$, so each branch preserves its pretrained transformation under the identity shared operator.

Such a design enforces that the subspace transformations in different modalities are parameterized by a common operator, while preserving modality-specific representation bases. As a result, the transformations across branches are coupled during optimization through shared parameters, enabling consistent updates within their respective subspaces.
{\color{black}
\begin{algorithm}[tb]
    \caption{Implementation of DROS}
    \label{alg:road_standard}
    \textbf{Input}: Pretrained weights $\{W_k^{\mathrm{pre}}\in\mathbb{R}^{d_{\mathrm{out},k}\times d_{\mathrm{in},k}}\}_{k=1}^{n}$; Representative training subset $\mathcal{D}_{\mathrm{sub}}$ with $N=\lvert\mathcal{D}_{\mathrm{sub}}\rvert$; Rank $r$. \\
    \textbf{Output}: Fine-tuned weights $\{W_k^{\mathrm{final}}\}_{k=1}^{n}$.

    \begin{algorithmic}[1]
        \STATE // \textit{Phase 1: Layer-wise activation statistics}
        \STATE Use $\mathcal{D}_{\mathrm{sub}}$ without labels or random augmentation.
        \FOR{$k=1,\ldots,n$}
            \STATE Collect $H_{k,i}\in\mathbb{R}^{d_{\mathrm{in},k}\times M_k}$ for $i=1,\ldots,N$ with the backbone frozen.
            \STATE $C_k \leftarrow \frac{1}{N}\sum_{i=1}^{N}H_{k,i}H_{k,i}^\top$.
        \ENDFOR

        \STATE // \textit{Phase 2: Covariance-conditioned subspace construction}
        \FOR{$k=1,\ldots,n$}
            \STATE $W_k^{\mathrm{rw}} \leftarrow W_k^{\mathrm{pre}}C_k = U_k\Sigma_kV_k^\top$.
            \STATE $B_{k,0} \leftarrow U_k[:, -r:]\sqrt{\Sigma_k[-r:]}$.
            \STATE $A_{k,0} \leftarrow \sqrt{\Sigma_k[-r:]}\left(V_k^\top C_k^{-1}\right)[-r:, :]$.
            \STATE $B_k\leftarrow B_{k,0}$, $A_k\leftarrow A_{k,0}$, $W_k^{\mathrm{res}}\leftarrow W_k^{\mathrm{pre}}-B_{k,0}A_{k,0}$.
            \STATE $Q_k\leftarrow0$, $\alpha_k\leftarrow\mathbf{1}_r$, $\beta_k\leftarrow\mathbf{1}_r$.
        \ENDFOR

        \STATE // \textit{Phase 3: Relaxed orthogonal adaptation}
        \WHILE{training}
            \FOR{$k=1,\ldots,n$}
                \STATE Reconstruct $O_k$ from $Q_k$.
                \STATE $\mathcal{T}_k\leftarrow\operatorname{diag}(\alpha_k)O_k\operatorname{diag}(\beta_k)$.
                \STATE $W_k^{\mathrm{side}}\leftarrow B_k\mathcal{T}_kA_k$.
            \ENDFOR
            \STATE Update $\{Q_k,\alpha_k,\beta_k,A_k,B_k\}_{k=1}^{n}$ by minimizing $\mathcal{L}_{\mathrm{task}}$.
        \ENDWHILE

        \STATE // \textit{Phase 4: Inference-efficient integration}
        \FOR{$k=1,\ldots,n$}
            \STATE Recompute $O_k$, $\mathcal{T}_k$, and $W_k^{\mathrm{side}}\leftarrow B_k\mathcal{T}_kA_k$ from the final parameters.
        \ENDFOR
        \STATE \textbf{return} $\{W_k^{\mathrm{final}}=W_k^{\mathrm{res}}+W_k^{\mathrm{side}}\}_{k=1}^{n}$
    \end{algorithmic}
\end{algorithm}
}
\section{Experiments}
\subsection{Experimental setup}
A comprehensive set of experiments is conducted to evaluate the effectiveness of DROS and MM-DROS on multiple remote sensing benchmarks, including image classification, semantic segmentation, domain-generalized semantic segmentation (DGSS), and multimodal segmentation. In addition, ablation studies are performed to analyze the contributions of individual components in terms of generalization and representational capacity.
\subsubsection{Baselines} DROS is evaluated on two representative foundation models: DINOv2-L\cite{dinov2}, a vision foundation model pretrained via self-supervised contrastive learning, and MARS\cite{mars}, a remote sensing foundation model pretrained using MAE. Comparisons are conducted against the following fine-tuning strategies:
\begin{itemize}
\item Full Fine-tuning (FT), which updates all model parameters during training.
\item Frozen (FZ), where the backbone is frozen and only task-specific heads are optimized.
\item LoRA\cite{lora}, which freezes pretrained weights and adapts the model via two low-rank matrices.
\item PiSSA\cite{pissa}, which improves LoRA by initializing updates along principal singular directions.
\item Earth-Adapter\cite{earth-adapter}, a remote sensing specific adapter that combines frequency decomposition with a mixture of adapter experts.
\item VPT\cite{vpt}, which prepends learnable prompts to frozen vision transformers at one or every layer.
\end{itemize}
\subsubsection{Datasets} 
Experiments are conducted on eleven representative remote sensing datasets, covering image classification, semantic segmentation, domain-generalized semantic segmentation (DGSS), and multimodal segmentation. For image classification, EuroSAT\cite{eurosat}, RESISC45\cite{resisc}, and UCM\cite{ucm} are adopted, using official splits provided by TorchGeo v0.6.2\cite{torchgeo}. For semantic segmentation, three tasks from GEO-Bench\cite{geo-bench} are selected, and the predefined data splits are strictly followed. For DGSS, following previous work\cite{earth-adapter}, experiments are conducted on LoveDA\cite{loveda}, ISPRS Potsdam\cite{p&v}, and Vaihingen\cite{p&v} under four cross-domain settings: Potsdam $\rightarrow$ Vaihingen (P2V), Vaihingen $\rightarrow$ Potsdam (V2P), Rural $\rightarrow$ Urban (R2U), and Urban $\rightarrow$ Rural (U2R). For multimodal segmentation, BRIGHT\cite{bright} and WHU-OPT-SAR\cite{whu-opt-sar} are adopted, and the predefined data splits are strictly followed. Additional dataset details are provided in Tab. \ref{tab:dataset_main}.
\subsubsection{Implementation details}
DROS is attached to every linear transformation in each Transformer block of the backbone. The adapted modules comprise the query, key, and value projections, the attention output projection, and the two feed forward network projections for channel expansion and contraction. Architectures that implement the three attention inputs through a fused QKV projection treat it as one adapted linear module. The embedding stem, normalization layers, and task-specific classifier or decoder are not equipped with DROS.
For classification tasks, MMClassification\cite{mmpretrain} is adopted as the training and evaluation framework, with a linear classifier appended to the backbone and trained for 30 epochs at $256\times256$ resolution and rank $r=1$. For semantic segmentation and DGSS, MMSegmentation\cite{mmseg} is adopted with Mask2Former\cite{mask2former} as the decoder, trained for 20,000 iterations, preserving the original image resolutions for segmentation and cropping $512\times512$ patches for DGSS, following Earth-Adapter\cite{earth-adapter}. To construct the domain-aware subspace, 10\% of the training data are randomly sampled from the downstream training split as representative unlabeled samples, which are used without stochastic data augmentation; only the deterministic preprocessing required by the corresponding foundation model is applied, and all backbone parameters remain frozen during activation collection. In DGSS, these samples are drawn exclusively from the source-domain training split. The standard, non-truncated SVD is applied to the covariance-conditioned weight matrix following the routine used in PiSSA\cite{pissa}; the inverse term in the Cayley transform\cite{cayley} is evaluated with a fifth-order truncated Neumann series following OFT-v2\cite{oftv2}. All experiments use the AdamW\cite{adamw} optimizer with batch size 4, a weight decay of 0.05, betas of 0.9 and 0.999, and eps of $1\times10^{-8}$, on a single NVIDIA A800 GPU with 40\,GB of memory in full FP32 precision, which is used rather than mixed precision to ensure the numerical stability of the covariance computation, SVD, and Neumann approximation, and to keep all methods under identical precision. The learning rates are $1\times10^{-5}$ for the backbone, $1\times10^{-4}$ for the decoder, and $1\times10^{-4}$ for PEFT parameters, following Earth-Adapter\cite{earth-adapter}. For full fine-tuning, the backbone and head learning rates are tuned separately by grid search: $1\times10^{-5}$ and $1\times10^{-3}$ for classification, and $1\times10^{-6}$ and $1\times10^{-4}$ for semantic segmentation and DGSS. The learning rate follows a polynomial decay schedule by iteration, with power 0.9 and minimum 0. All baselines share the same data augmentation, optimizer, batch size, iterations, precision, and random seeds, and no class re-weighting or balancing is applied. For training, classification images are resized to $256\times256$; for segmentation and DGSS, RandomResize with scale $512$ and ratio range 0.5--2.0, RandomCrop with crop size $512$ and maximum category ratio 0.75, RandomFlip with probability 0.5, and PhotoMetricDistortion are applied. Top-1 accuracy and mIoU are reported as the primary metrics, and results are reported as the mean $\pm$ standard deviation over three random seeds.
\begin{table*}
\centering
\caption{Detailed statistics of the experimental datasets. For classification and GEO-Bench tasks, the official data splits provided by TorchGeo and GEO-Bench are adopted. For DGSS tasks, the number of cropped $512 \times 512$ patches used for training and evaluation is reported. The data source/sensor, covered region, and reported dataset year of each dataset are also given.}
\label{tab:dataset_main}
\small
\setlength{\tabcolsep}{2pt}
\begin{tabular}{lllccccc}
\toprule
\textbf{Task Type} & \textbf{Dataset} & {\textbf{\shortstack{Source (Region,\\ Reported Dataset Year)}}} & \textbf{Resolution} & \textbf{Classes} & \textbf{Train Size} & \textbf{Test Size} & {\textbf{Total Images}} \\ \midrule
\multirow{3}{*}{Classification}
& EuroSAT & {Sentinel-2, Europe (2017)} & 10m & 10 & 16,200 & 10,800 & 27,000 \\
& RESISC45 & {Google Earth, Global (2015)} & 0.3m & 45 & 18,900 & 6,300 & 25,200 \\
& UCM & {Aerial, California, US (2010)} & 0.3m & 21 & 1,260 & 420 & 1,680 \\ \midrule
\multirow{3}{*}{\shortstack{Semantic\\Segmentation}}
& Cashew & {Sentinel-2, Benin (GEO-Bench)} & 10m & 7 & 1,350 & 400 & 1,750 \\
& Chesap. & {RGBN, Chesapeake Bay, US (GEO-Bench)} & 1.0m & 7 & 3,000 & 1,000 & 4,000 \\
& Cattle & {RGB, New Zealand (GEO-Bench)} & 0.1m & 2 & 524 & 66 & 590 \\ \midrule
\multirow{4}{*}{\shortstack{Domain\\Generalization}}
& LoveDA(Urban) & {Google Earth, Nanjing/Wuhan, China (2016)} & 0.3m & 7 & 4,625 & 2,709 & 7,344\\
& LoveDA(Rural) & {Google Earth, Nanjing/Wuhan, China (2016)} & 0.3m & 7 & 5,465 & 3,969 & 9,434 \\
& Potsdam & {Aerial, Germany, ISPRS (2013)} & 0.05m & 6 & 3,457 & 2,017 & 5,474 \\
& Vaihingen & {Aerial, Germany, ISPRS (2013)} & 0.05m & 6 & 345 & 399 & 744 \\ \midrule
\multirow{2}{*}{\shortstack{Multimodal\\Segmentation}}
& BRIGHT & {Optical+SAR, Global (2021--2023)} & 0.3-1m & 2 & 8,916 & 2,644 & 11,560 \\
& WHU-OPT-SAR & {GF-1/GF-3, Hubei, China (2017)} & 5m & 8 & 7,040 & 1,760 & 8,800\\ \bottomrule
\end{tabular}%

\end{table*}
\subsection{Experimental Result Analysis}
All main performance results are reported as the mean $\pm$ standard deviation over three independent runs. Because the number of runs is limited and individual margins are modest in several settings, conclusions are not based on per-cell significance tests; instead, the consistency of DROS's improvements across datasets and backbones is emphasized, and each gain is reported relative to the strongest competing PEFT method in the corresponding setting. As a quantitative check of this consistency, a two-sided sign test across the dataset--backbone settings of Tabs. \ref{tab:cls_results}--\ref{tab:dg_results_final} shows that DROS improves over PiSSA in 19 of 20 settings and over LoRA in 18 of 20 settings, with $p\approx 4\times10^{-5}$ and $p\approx 4\times10^{-4}$ respectively, providing statistical evidence for the consistency of the improvement. Three seeds follow common practice for large-foundation-model fine-tuning, where additional runs incur substantial computational cost without altering the conclusions.
\subsubsection{Image classification} 
As summarized in Tab. \ref{tab:cls_results}, DROS attains the highest average accuracy on both MARS (97.32\%) and DINOv2-L (96.85\%), and it improves over LoRA and PiSSA in average accuracy on both backbones under identical parameter constraints. On MARS, DROS achieves the best accuracy on EuroSAT and UCM, at 96.63\% and 98.89\%; on DINOv2-L, it achieves the best UCM accuracy of 98.49\% and the best average accuracy of 96.85\%. The individual margins over the strongest PEFT competitors are small, mostly below 0.5 percentage points, consistent with the near-saturated 95--99\% accuracy range of these benchmarks; the advantage is therefore most visible in the average metric, where the consistent though modest edge of DROS accumulates. This indicates that scene classification is less sensitive to subspace adaptation, as global semantic representations are largely preserved in pretrained backbones. Nevertheless, by leveraging domain-aware statistics and selectively activating informative components, DROS maintains a favorable robustness--efficiency trade-off in mitigating domain discrepancies.

\begin{figure}
    \centering
    \includegraphics[width=\columnwidth]{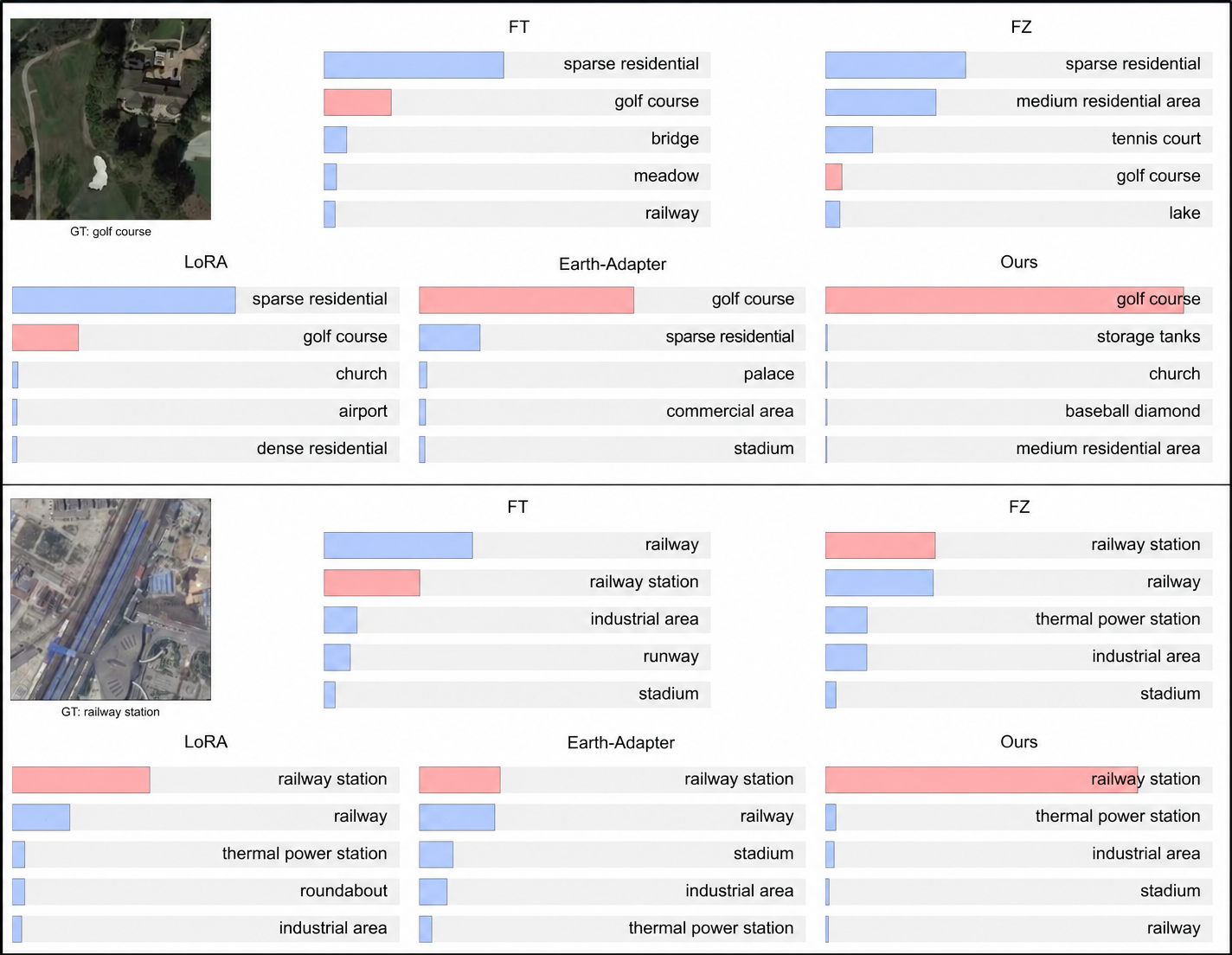}
    \caption{Qualitative comparison of Top-5 classification predictions on RESISC45 using DINOv2-L. Red bars identify the ground-truth category, and Ours denotes DROS.}
    \label{fig:top5}
\end{figure}

Fig. \ref{fig:top5} compares DROS with FT, FZ, LoRA, and Earth-Adapter on two RESISC45 examples using DINOv2-L. For the golf course image, all four baselines rank sparse residential above the ground-truth class, whereas DROS assigns the highest confidence to golf course. For the railway station image, FZ, LoRA, and Earth-Adapter recover the correct top-ranked class, but DROS assigns it a visibly larger confidence margin. These examples show that DROS improves both the ranking and confidence separation of the ground-truth category.
\begin{table*}[!tbp]
\centering
\caption{Comparative results on classification tasks using MARS and DINOv2-L backbones. {\color{red}Red} and {\color{blue}blue} indicate the best and second-best results, respectively.}
\label{tab:cls_results}
\setlength{\tabcolsep}{0pt}
\begin{tabular*}{\textwidth}{@{\extracolsep{\fill}} l c cccc c cccc}
\toprule
\multirow{2}{*}{\shortstack{\textbf{Method}}} & \multirow{2}{*}{\shortstack{\textbf{\#Params}}} & \multicolumn{4}{c}{\textbf{MARS}} & \multirow{2}{*}{\shortstack{\textbf{\#Params}}} & \multicolumn{4}{c}{\textbf{DINOv2-L}} \\ 
\cmidrule(lr){3-6} \cmidrule(lr){8-11}
& & \textbf{EuroSAT} & \textbf{UCM} & \textbf{RESISC45} & \textbf{Avg} & & \textbf{EuroSAT} & \textbf{UCM} & \textbf{RESISC45} & \textbf{Avg} \\ 
\midrule
FT    & 89.00M & 96.47 $\pm$ 0.31 & 98.57 $\pm$ 0.00 & {\color[HTML]{FE0000} 96.85 $\pm$ 0.17} & 97.30 & 304.20M & 96.63 $\pm$ 0.12 & {\color[HTML]{3531FF} 98.41 $\pm$ 0.69} & 95.25 $\pm$ 0.18 & 96.76 \\
FZ    & 0.00M  & 86.83 $\pm$ 0.17 & 93.17 $\pm$ 0.11 & 87.84 $\pm$ 0.04 & 89.28 & 0.00M & 86.13 $\pm$ 0.49 & 90.00 $\pm$ 1.04 & 84.23 $\pm$ 0.12 & 86.79 \\
LoRA  & 0.28M  & 95.77 $\pm$ 0.29 & 98.73 $\pm$ 0.36 & 96.49 $\pm$ 0.10 & 97.00 & 0.59M & {\color[HTML]{FE0000} 96.83 $\pm$ 0.31} & 98.10 $\pm$ 0.41 & 95.33 $\pm$ 0.20 & 96.75 \\
{Earth-Adapter} & 1.55M & {\color[HTML]{3531FF} 96.50 $\pm$ 0.10} & {\color[HTML]{3531FF} 98.69 $\pm$ 0.11} & {\color[HTML]{3531FF} 96.75 $\pm$ 0.05} & {\color[HTML]{3531FF} 97.31} & 9.59M & 96.37 $\pm$ 0.12 & 98.41 $\pm$ 1.10 & {\color[HTML]{FE0000} 95.62 $\pm$ 0.19} & {\color[HTML]{3531FF} 96.80} \\
{VPT}   & 0.24M & 86.97 $\pm$ 0.06 & 92.54 $\pm$ 0.14 & 87.80 $\pm$ 0.06 & 89.10 & 0.49M & 96.43 $\pm$ 0.25 & 97.86 $\pm$ 0.24 & 95.12 $\pm$ 0.06 & 96.47 \\
PiSSA & 0.28M & 95.33 $\pm$ 0.32 & 98.41 $\pm$ 0.36 & 96.45 $\pm$ 0.06 & 96.73 & 0.59M & 96.60 $\pm$ 0.30 & 98.10 $\pm$ 0.63 & 95.10 $\pm$ 0.37 & 96.60 \\
DROS  & 0.28M & {\color[HTML]{FE0000} 96.63 $\pm$ 0.06} & {\color[HTML]{FE0000} 98.89 $\pm$ 0.11} & 96.44 $\pm$ 0.15 & {\color[HTML]{FE0000} 97.32} & 0.59M & {\color[HTML]{3531FF} 96.70 $\pm$ 0.66} & {\color[HTML]{FE0000} 98.49 $\pm$ 0.36} & {\color[HTML]{3531FF} 95.37 $\pm$ 0.10} & {\color[HTML]{FE0000} 96.85} \\
\bottomrule
\end{tabular*}%

\end{table*}
\subsubsection{Semantic segmentation} 
As shown in Tab. \ref{tab:segmentation_results}, on DINOv2-L DROS achieves the highest mIoU on all three datasets and in the average, at 75.57\%, surpassing all baselines including full fine-tuning. Under MARS, full fine-tuning remains the strongest overall, with the best results on Cashew and Chesap.\ and the best average of 79.17\%, yet among parameter-efficient methods DROS is again the best on every dataset and in the average, at 78.23\%. The gains over the strongest PEFT competitors are generally larger than in classification, and are especially pronounced on Cashew under DINOv2-L, where the margin over PiSSA reaches 2.85 points. The larger margins on segmentation indicate that pixel-level prediction is more sensitive to domain-aware subspace adaptation than scene classification, as dense tasks require precise modeling of local spatial variations. By selectively activating informative components while suppressing less relevant directions, DROS effectively preserves fine-grained spatial structures without compromising global consistency, leading to improved segmentation accuracy across domains.

\begin{figure}
    \centering
    \includegraphics[width=\columnwidth]{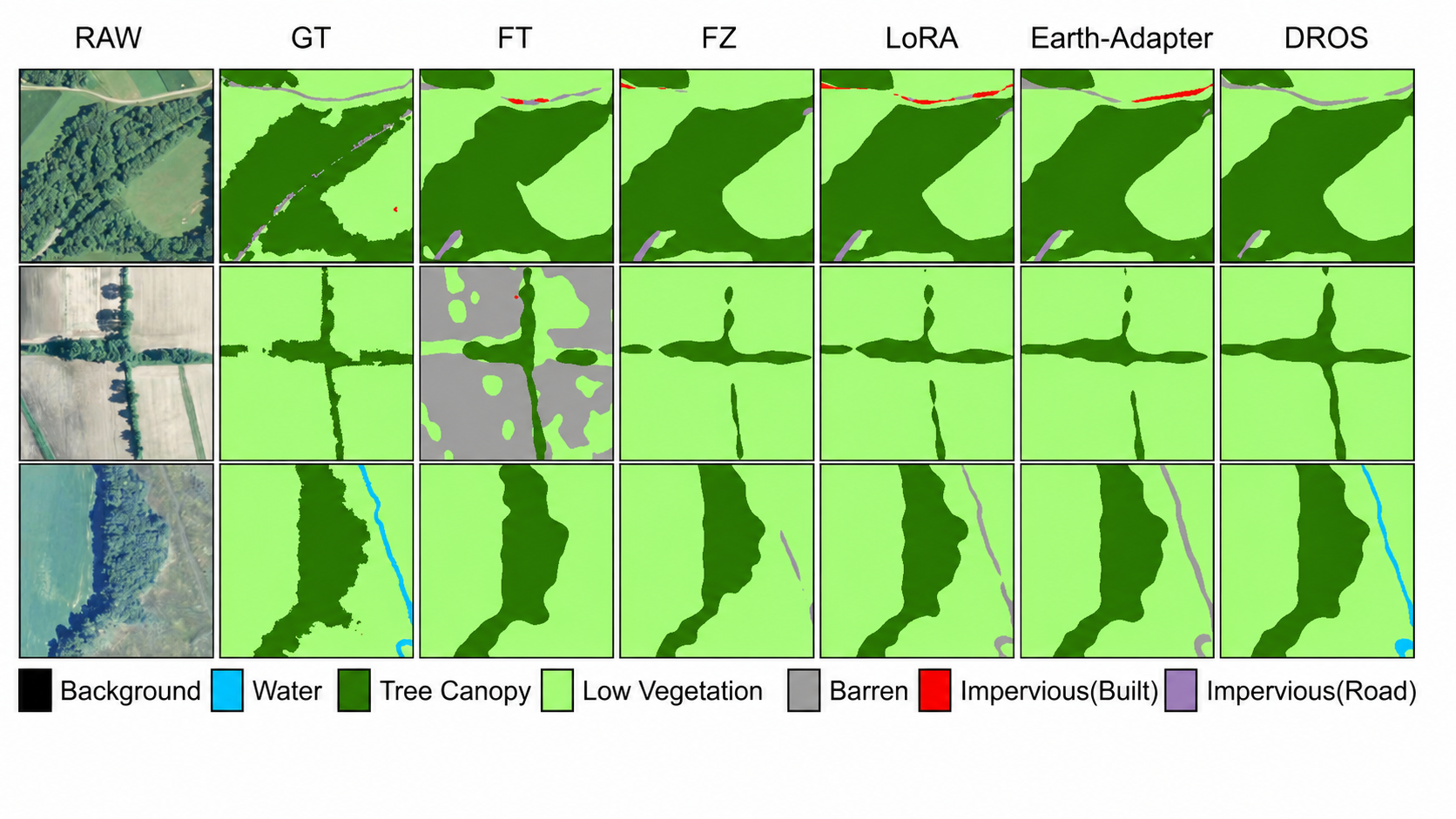}
    \caption{Qualitative comparison on Chesapeake semantic segmentation using DINOv2-L. }
    \label{fig:ches}
\end{figure}

Fig. \ref{fig:ches} compares DROS with FT, FZ, LoRA, and Earth-Adapter on the Chesapeake dataset using DINOv2-L. Full fine-tuning produces large fragmented errors in the second example. FZ, LoRA, and Earth-Adapter recover the dominant land-cover regions but miss or distort fine structures, most clearly the narrow water channel in the third example. DROS follows the ground-truth geometry more closely and preserves the channel continuously, demonstrating improved boundary and topology preservation.
\begin{table*}[!tbp]
\centering
\caption{Comparative results on segmentation tasks using MARS and DINOv2-L backbones. For clarity, dataset names are abbreviated: \textbf{Cashew}: m-cashew-plant; \textbf{Chesap.}: m-chesapeake; \textbf{Cattle}: m-nz-cattle. {\color[HTML]{FE0000} Red} and {\color[HTML]{3531FF} blue} indicate the best and second-best performance, respectively.}
\label{tab:segmentation_results}
\setlength{\tabcolsep}{0pt}
\begin{tabular*}{\textwidth}{@{\extracolsep{\fill}} l c cccc c cccc}
\toprule
\multirow{2}{*}{\shortstack{\textbf{Method}}} & \multirow{2}{*}{\shortstack{\textbf{\#Params}}} & \multicolumn{4}{c}{\textbf{MARS}} & \multirow{2}{*}{\shortstack{\textbf{\#Params}}} & \multicolumn{4}{c}{\textbf{DINOv2-L}} \\ 
\cmidrule(lr){3-6} \cmidrule(lr){8-11}
& & \textbf{Cashew} & \textbf{Chesap.} & \textbf{Cattle} & \textbf{Avg} & & \textbf{Cashew} & \textbf{Chesap.} & \textbf{Cattle} & \textbf{Avg} \\ 
\midrule
FT    & 89.00M & {\color[HTML]{FE0000} 78.30 $\pm$ 0.20} & {\color[HTML]{FE0000} 75.53 $\pm$ 0.75} & 83.69 $\pm$ 0.19 & {\color[HTML]{FE0000} 79.17} & 304.20M & 62.65 $\pm$ 1.69 & 74.59 $\pm$ 0.18 & {\color[HTML]{3531FF} 83.18 $\pm$ 0.54} & 73.47 \\
FZ    & 0.00M & 69.05 $\pm$ 1.06 & 74.15 $\pm$ 0.37 & 84.02 $\pm$ 0.48 & 75.74 & 0.00M & 55.68 $\pm$ 4.26 & 73.35 $\pm$ 0.16 & 82.55 $\pm$ 0.14 & 70.53 \\
LoRA  & 3.03M & 74.82 $\pm$ 1.13 & 75.25 $\pm$ 0.17 & 83.82 $\pm$ 0.11 & 77.96 & 6.21M & 63.53 $\pm$ 1.62 & {\color[HTML]{3531FF} 75.13 $\pm$ 0.18} & 83.10 $\pm$ 0.31 & 73.92 \\
{Earth-Adapter} & 1.55M & 70.29 $\pm$ 0.80 & 74.56 $\pm$ 0.41 & 83.57 $\pm$ 0.10 & 76.14 & 9.59M & 58.98 $\pm$ 1.80 & 74.04 $\pm$ 0.16 & 82.89 $\pm$ 0.31 & 71.97 \\
{VPT}   & 3.08M & 69.83 $\pm$ 0.30 & 74.59 $\pm$ 0.24 & {\color[HTML]{3531FF} 84.03 $\pm$ 0.56} & 76.15 & 6.29M & 50.51 $\pm$ 1.46 & 74.22 $\pm$ 0.26 & 82.94 $\pm$ 0.55 & 69.22 \\
PiSSA & 3.03M & 72.59 $\pm$ 1.84 & 74.60 $\pm$ 0.06 & 83.75 $\pm$ 0.17 & 76.98 & 6.21M & {\color[HTML]{3531FF} 64.62 $\pm$ 0.61} & 75.01 $\pm$ 0.36 & 83.16 $\pm$ 0.08 & {\color[HTML]{3531FF} 74.26} \\
DROS  & 3.04M & {\color[HTML]{3531FF} 75.31 $\pm$ 0.30} & {\color[HTML]{3531FF} 75.31 $\pm$ 0.31} & {\color[HTML]{FE0000} 84.07 $\pm$ 0.45} & {\color[HTML]{3531FF} 78.23} & 6.22M & {\color[HTML]{FE0000} 67.47 $\pm$ 0.75} & {\color[HTML]{FE0000} 75.58 $\pm$ 0.13} & {\color[HTML]{FE0000} 83.67 $\pm$ 0.27} & {\color[HTML]{FE0000} 75.57} \\
\bottomrule
\end{tabular*}%

\end{table*}
\subsubsection{Domain generalization semantic segmentation}
For DGSS, the covariance statistics used for subspace construction are estimated only from the source-domain training split; no unlabeled or labeled samples from the unseen target domain are used during training or subspace construction. As shown in Tab. \ref{tab:dg_results_final}, DROS achieves the best average mIoU across all DGSS settings on both backbones, 44.81\% on MARS and 55.32\% on DINOv2-L, and it attains the best result in seven of the eight setting--backbone combinations---all four settings under MARS, and R2U, P2V, and V2P under DINOv2-L, where Earth-Adapter is marginally ahead on U2R with 45.99\% against 45.77\%. DROS improves over LoRA and PiSSA on every setting, and the gains are generally larger under severe domain shifts, such as cross-city transfer in P2V and V2P and cross-scene transfer in U2R and R2U. This highlights that DGSS is highly sensitive to domain discrepancies, as it requires learning domain-invariant representations across diverse geographical and environmental conditions. By leveraging domain-aware covariance statistics and selectively activating informative subspace components, DROS effectively identifies task-relevant directions for generalization, leading to superior adaptability to unseen domains and improved pixel-level accuracy under substantial distribution shifts.

\begin{figure}
    \centering
    \includegraphics[width=\columnwidth]{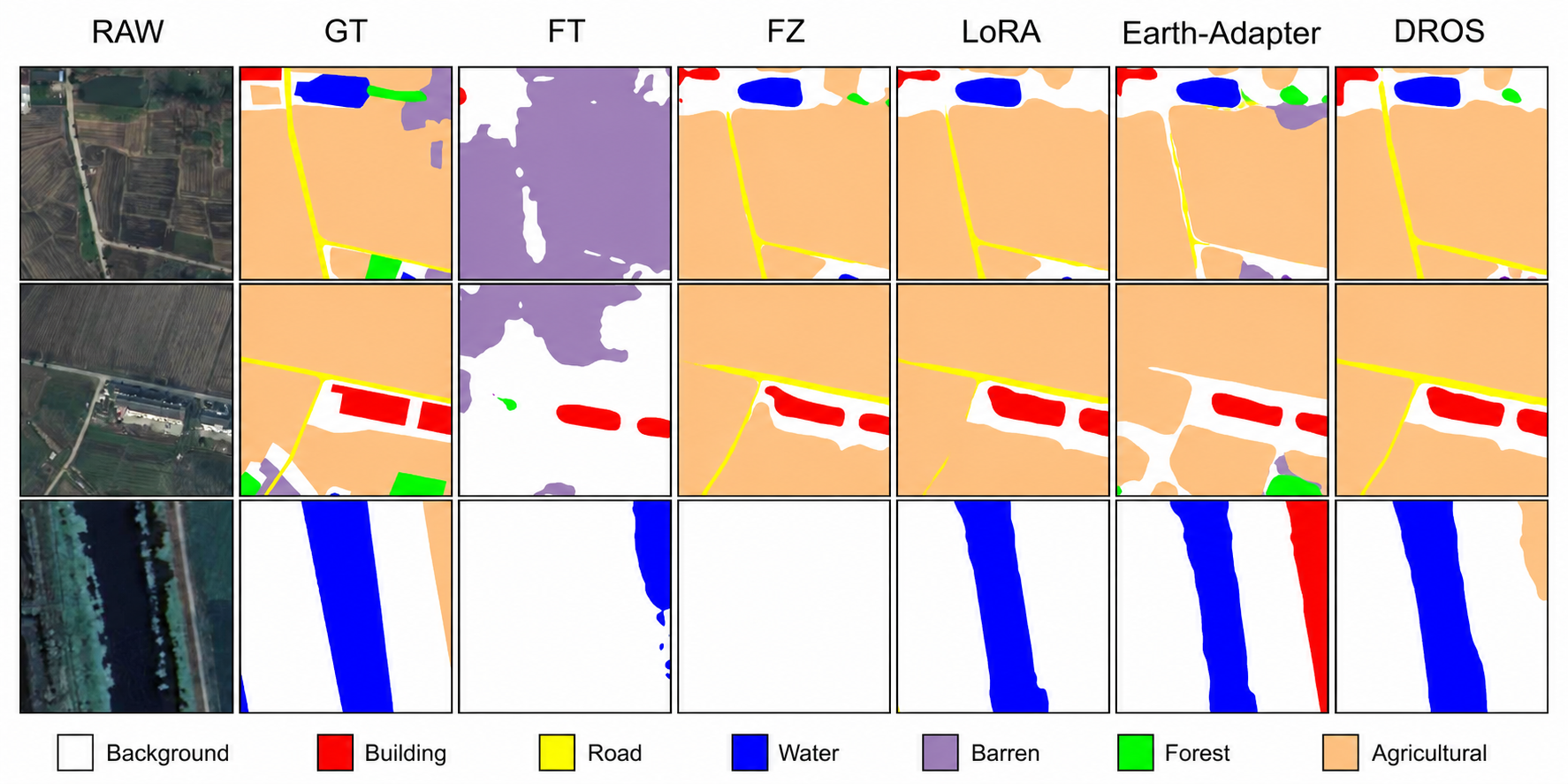}
    \caption{Qualitative comparison on the LoveDA U2R setting using DINOv2-L. }
    \label{fig:u2r}
\end{figure}

Fig. \ref{fig:u2r} compares DROS with FT, FZ, LoRA, and Earth-Adapter on the U2R setting of LoveDA using DINOv2-L. Under the pronounced urban-to-rural domain shift, FT exhibits severe semantic drift, while FZ and LoRA omit important classes or distort object boundaries. Earth-Adapter preserves more of the scene structure but introduces background gaps and class confusion. In the third example, Earth-Adapter recovers the water region but adds a spurious building strip, whereas DROS preserves the continuous river without this error. Across the three examples, DROS more closely matches the agricultural boundaries, roads, and water topology in the ground truth.
\begin{table*}[!tbp]
\centering
\caption{Domain generalization results (mIoU \%) on MARS and DINOv2-L backbones. Tasks include cross-city (\textbf{P2V}, \textbf{V2P}) and cross-scene (\textbf{U2R}, \textbf{R2U}) transfers. {\color[HTML]{FE0000} Red} and {\color[HTML]{3531FF} blue} indicate the best and second-best results, respectively.}
\label{tab:dg_results_final}
\setlength{\tabcolsep}{0pt}
%
\begin{tabular*}{\textwidth}{@{\extracolsep{\fill}} l c ccccc c ccccc}
\toprule
\multirow{2}{*}{\shortstack{\textbf{Method}}} & \multirow{2}{*}{\shortstack{\textbf{\#Params}}} & \multicolumn{5}{c}{\textbf{MARS}} & \multirow{2}{*}{\shortstack{\textbf{\#Params}}} & \multicolumn{5}{c}{\textbf{DINOv2-L}} \\ 
\cmidrule(lr){3-7} \cmidrule(lr){9-13}
& & \textbf{U2R} & \textbf{R2U} & \textbf{P2V} & \textbf{V2P} & \textbf{Avg} & & \textbf{U2R} & \textbf{R2U} & \textbf{P2V} & \textbf{V2P} & \textbf{Avg} \\ 
\midrule
FT    & 89.00M & 38.82 $\pm$ 0.76 & 52.90 $\pm$ 0.54 & 41.22 $\pm$ 0.56 & {\color[HTML]{3531FF} 40.78 $\pm$ 0.94} & 43.43 & 304.20M & 45.55 $\pm$ 0.82 & 58.38 $\pm$ 0.39 & 60.72 $\pm$ 1.31 & 55.57 $\pm$ 0.88 & {\color[HTML]{3531FF} 55.05} \\
FZ    & 0.00M & 36.69 $\pm$ 0.62 & 50.87 $\pm$ 1.16 & 37.32 $\pm$ 0.79 & 32.79 $\pm$ 1.15 & 39.42 & 0.00M & 45.23 $\pm$ 1.40 & 57.37 $\pm$ 0.04 & 55.74 $\pm$ 1.14 & 49.93 $\pm$ 1.28 & 52.07 \\
LoRA  & 3.03M & 39.67 $\pm$ 0.60 & {\color[HTML]{3531FF} 53.84 $\pm$ 0.08} & 41.32 $\pm$ 1.63 & 39.39 $\pm$ 0.70 & {\color[HTML]{3531FF} 43.56} & 6.21M & 45.27 $\pm$ 1.11 & 58.02 $\pm$ 0.30 & {\color[HTML]{3531FF} 60.76 $\pm$ 0.65} & 54.14 $\pm$ 0.86 & 54.55 \\
{Earth-Adapter} & 1.55M & {\color[HTML]{3531FF} 40.33 $\pm$ 1.25} & 52.34 $\pm$ 0.70 & 39.31 $\pm$ 0.93 & 40.59 $\pm$ 0.53 & 43.14 & 9.59M & {\color[HTML]{FE0000} 45.99 $\pm$ 0.94} & 57.72 $\pm$ 0.35 & 59.49 $\pm$ 1.00 & 55.66 $\pm$ 1.51 & 54.72 \\
{VPT}   & 0.77M & 36.78 $\pm$ 1.69 & 51.47 $\pm$ 1.17 & 37.51 $\pm$ 0.28 & 35.51 $\pm$ 0.19 & 40.32 & 1.57M & 43.51 $\pm$ 1.99 & 57.46 $\pm$ 0.10 & 58.27 $\pm$ 0.49 & {\color[HTML]{3531FF} 55.75 $\pm$ 0.93} & 53.75 \\
PiSSA & 3.03M & 38.57 $\pm$ 1.32 & 53.43 $\pm$ 0.42 & {\color[HTML]{3531FF} 42.03 $\pm$ 0.77} & 39.91 $\pm$ 0.53 & 43.48 & 6.21M & 45.07 $\pm$ 1.23 & {\color[HTML]{3531FF} 58.43 $\pm$ 0.19} & 59.24 $\pm$ 1.90 & 55.32 $\pm$ 0.68 & 54.52 \\
DROS  & 3.04M & {\color[HTML]{FE0000} 41.02 $\pm$ 1.63} & {\color[HTML]{FE0000} 54.05 $\pm$ 0.10} & {\color[HTML]{FE0000} 42.30 $\pm$ 0.92} & {\color[HTML]{FE0000} 41.87 $\pm$ 0.69} & {\color[HTML]{FE0000} 44.81} & 6.22M & {\color[HTML]{3531FF} 45.77 $\pm$ 0.61} & {\color[HTML]{FE0000} 58.47 $\pm$ 0.37} & {\color[HTML]{FE0000} 60.91 $\pm$ 0.84} & {\color[HTML]{FE0000} 56.14 $\pm$ 1.09} & {\color[HTML]{FE0000} 55.32} \\
\bottomrule
\end{tabular*}%

\end{table*}
\subsubsection{Multimodal semantic segmentation}
As shown in Tab. \ref{tab:mm_results}, MM-DROS outperforms the independent-branch DROS on both backbones, demonstrating the effectiveness of cross-modal parameter sharing. On DINOv2-L, MM-DROS achieves an average mIoU of 63.09\%, surpassing the independent-branch DROS at 62.97\% by 0.12 points; on MARS, it improves from 61.98\% to 62.12\%. The gains are small but consistent: MM-DROS improves on both BRIGHT and WHU-OPT-SAR under each backbone. This improvement validates the proposed shared orthogonal operator, which enables coordinated subspace transformations across modalities. By coupling modality-specific representations through a shared transformation structure, MM-DROS effectively aligns heterogeneous feature spaces and facilitates cross-modal gradient interaction, leading to more coherent and discriminative segmentation while reducing parameter redundancy.

\begin{figure}
    \centering
    \includegraphics[width=\columnwidth]{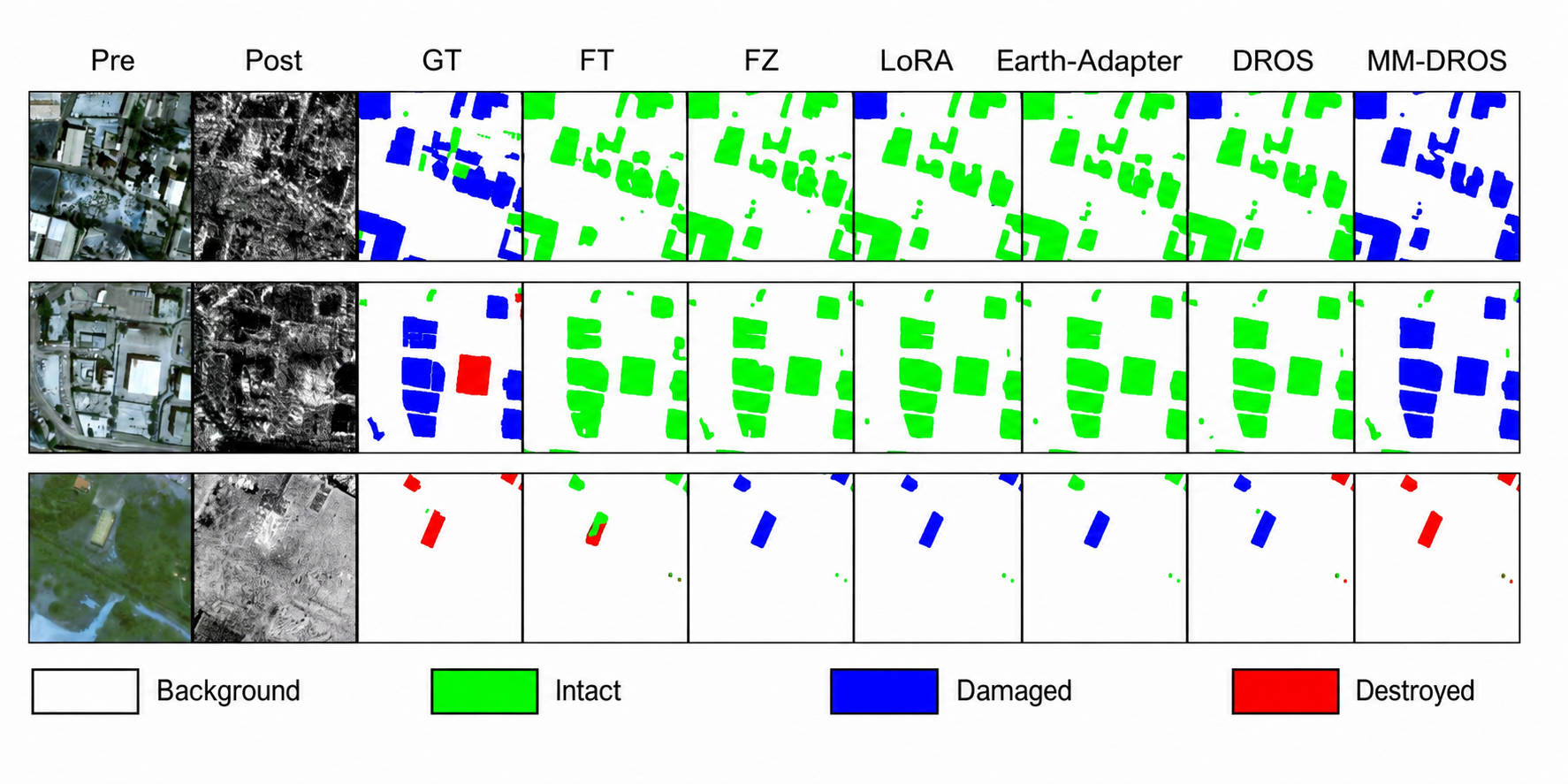}
    \caption{Qualitative comparison on BRIGHT multimodal segmentation using DINOv2-L. }
    \label{fig:bright}
\end{figure}

Fig. \ref{fig:bright} compares MM-DROS with FT, FZ, LoRA, Earth-Adapter, and the independent-branch DROS on BRIGHT using DINOv2-L. The first two examples show persistent confusion between damaged and intact buildings for FT, FZ, LoRA, Earth-Adapter, and DROS, whereas MM-DROS more closely matches the affected regions in the ground truth. In the third example, FZ, LoRA, Earth-Adapter, and DROS assign the damaged label to the main destroyed building, while FT assigns most of this building to the intact class. MM-DROS recovers both the main destroyed region and the smaller destroyed objects at their ground-truth locations. This comparison supports the role of the shared operator in coordinating optical and SAR representations under multimodal distribution shifts.
\begin{table*}[!tbp]
\centering
\caption{Multimodal segmentation results (mIoU \%) on MARS and DINOv2-L backbones. For clarity, dataset names are abbreviated: \textbf{WHU}: WHU-OPT-SAR. {\color[HTML]{FE0000} Red} and {\color[HTML]{3531FF} blue} indicate the best and second-best results, respectively.}
\label{tab:mm_results}
\setlength{\tabcolsep}{0pt}
\begin{tabular*}{\textwidth}{@{\extracolsep{\fill}} l c ccc c ccc}
\toprule
\multirow{2}{*}{\shortstack{\textbf{Method}}} & \multirow{2}{*}{\shortstack{\textbf{\#Params}}} & \multicolumn{3}{c}{\textbf{MARS}} & \multirow{2}{*}{\shortstack{\textbf{\#Params}}} & \multicolumn{3}{c}{\textbf{DINOv2-L}} \\ 
\cmidrule(lr){3-5} \cmidrule(lr){7-9}
& & \textbf{BRIGHT} & \textbf{WHU} & \textbf{Avg} & & \textbf{BRIGHT} & \textbf{WHU} & \textbf{Avg} \\ 
\midrule
FT    & 178.00M & {\color[HTML]{3531FF} 63.78 $\pm$ 0.36} & 58.59 $\pm$ 0.59 & 61.19 & 608.40M & 65.28 $\pm$ 0.62 & 60.36 $\pm$ 0.05 & 62.82 \\
FZ    & 0.00M & 55.22 $\pm$ 1.35 & 59.22 $\pm$ 0.08 & 57.22 & 0.00M & 62.08 $\pm$ 1.69 & 59.80 $\pm$ 0.07 & 60.94 \\
LoRA  & 6.06M & 63.29 $\pm$ 0.61 & 59.89 $\pm$ 0.27 & 61.59 & 12.42M & 64.88 $\pm$ 1.48 & 60.32 $\pm$ 0.33 & 62.60 \\
{Earth-Adapter} & 3.11M & 62.62 $\pm$ 1.61 & 59.79 $\pm$ 0.06 & 61.20 & 19.18M & 65.08 $\pm$ 0.87 & 60.32 $\pm$ 0.15 & 62.70 \\
{VPT}   & 6.16M & 54.65 $\pm$ 1.06 & 59.14 $\pm$ 0.13 & 56.89 & 12.58M & 64.11 $\pm$ 1.20 & 60.25 $\pm$ 0.19 & 62.18 \\
PiSSA & 6.06M & 63.25 $\pm$ 0.71 & 60.11 $\pm$ 0.12 & 61.68 & 12.42M & 64.50 $\pm$ 0.42 & 60.32 $\pm$ 0.25 & 62.41 \\
DROS  & 6.08M & 63.73 $\pm$ 0.46 & {\color[HTML]{3531FF} 60.23 $\pm$ 0.22} & {\color[HTML]{3531FF} 61.98} & 12.44M & {\color[HTML]{3531FF} 65.41 $\pm$ 0.84} & {\color[HTML]{3531FF} 60.52 $\pm$ 0.15} & {\color[HTML]{3531FF} 62.97} \\
MM-DROS & 6.07M & {\color[HTML]{FE0000} 63.99 $\pm$ 0.12} & {\color[HTML]{FE0000} 60.24 $\pm$ 0.15} & {\color[HTML]{FE0000} 62.12} & 12.43M & {\color[HTML]{FE0000} 65.60 $\pm$ 0.84} & {\color[HTML]{FE0000} 60.58 $\pm$ 0.14} & {\color[HTML]{FE0000} 63.09} \\
\bottomrule
\end{tabular*}%

\end{table*}
\subsection{Ablation studies}
Ablation experiments are conducted on DINOv2-L to investigate the factors that may affect the performance of DROS.
\subsubsection{Component selection as mechanism verification: top-$r$ vs.\ bottom-$r$} The component-selection rule is directly verified by holding the covariance conditioning, backbone, rank, optimizer, and training protocol fixed and changing only whether the top-$r$ or bottom-$r$ components initialize the trainable branch. As shown in Tab.~\ref{tab:ablation_spectral}, top-$r$ adaptation obtains 25.21\% on RESISC45, 70.32\% on Chesap., and 40.97\% on U2R, whereas bottom-$r$ adaptation obtains 95.37\%, 75.58\%, and 45.77\%, respectively. Thus, modifying the dominant components causes a severe collapse in scene classification and consistent degradation on both dense-prediction tasks, while bottom-$r$ adaptation remains stable across the three settings. The degradation is particularly severe for classification, whose prediction depends directly on the global representation of the pretrained backbone, and is milder for dense tasks, where the decoder can partially compensate. These results support bottom-$r$ selection as a preservation-oriented inductive bias under the covariance-conditioned DROS formulation. They do not imply that principal-component adaptation is generally inferior or that minor components universally encode domain-specific information. Rather, PiSSA\cite{pissa} adapts the principal components of $W^{\mathrm{pre}}$ to improve early optimization, whereas DROS retains the dominant components of $W^{\mathrm{pre}}C$ in the frozen residual and adapts lower-energy directions to reduce perturbation of the strongest pretrained structure. The two choices therefore reflect different objectives rather than contradictory conclusions about singular-component tuning.
\begin{table}[!tbp]
\centering
\caption{Mechanism verification of spectral component selection on DINOv2-L. Adapting the top-$r$ principal components collapses classification accuracy and degrades dense tasks, whereas adapting the bottom-$r$ components remains stable.}
\label{tab:ablation_spectral}
\resizebox{\columnwidth}{!}{
\begin{tabular}{lcccc}
\toprule
\textbf{Component} & \textbf{RESISC45} & \textbf{Chesap.} & \textbf{U2R} & \textbf{Avg} \\ 
\midrule
Dominant (Top-$r$) & 25.21 $\pm$ 0.32 & 70.32 $\pm$ 0.59 & 40.97 $\pm$ 1.19 & 45.50 \\
\textbf{Minor (Bot-$r$)} & \textbf{95.37 $\pm$ 0.10} & \textbf{75.58 $\pm$ 0.13} & \textbf{45.77 $\pm$ 0.61} & \textbf{72.24} \\
\bottomrule
\end{tabular}%
}
\end{table}
\subsubsection{Impact of sample size on domain-aware covariance statistics} Tab. \ref{tab:ablation_subset} reports the effect of different proportions of training samples used for estimating the covariance matrix $C$. Performance is already competitive when 1\% of the training split is sampled and reaches its highest mean at 10\%. The differences among 1\%, 10\%, 50\%, and 100\% are modest and non-monotonic, and largely fall within run-to-run variability. The results are therefore interpreted as evidence that a small representative subset is sufficient, rather than attributing the differences to increased covariance-estimation noise.
\begin{table}[ht]
\centering
\caption{Ablation of training subset ratio for domain-aware covariance estimation on DINOv2-L. }
\label{tab:ablation_subset}
\resizebox{\columnwidth}{!}{
\begin{tabular}{lcccc}
\toprule
\textbf{Subset Ratio} & \textbf{RESISC45} & \textbf{Chesap.} & \textbf{U2R} & \textbf{Avg} \\ 
\midrule
1\%   & 94.80 $\pm$ 0.77 & 74.89 $\pm$ 0.45 & 45.48 $\pm$ 0.80 & 71.72 \\
\textbf{10\%}  & \textbf{95.37 $\pm$ 0.10} & \textbf{75.58 $\pm$ 0.13} & \textbf{45.77 $\pm$ 0.61} & \textbf{72.24} \\
50\%  & 94.74 $\pm$ 0.27 & 74.91 $\pm$ 0.20 & 45.59 $\pm$ 0.53 & 71.75 \\
100\% & 94.63 $\pm$ 0.59 & 74.96 $\pm$ 0.08 & 45.08 $\pm$ 0.89 & 71.56 \\
\bottomrule
\end{tabular}%
}
\end{table}
\subsubsection{Impact of rank r} 
\begin{figure}[!tbp]
    \centering
    \includegraphics[width=\columnwidth]{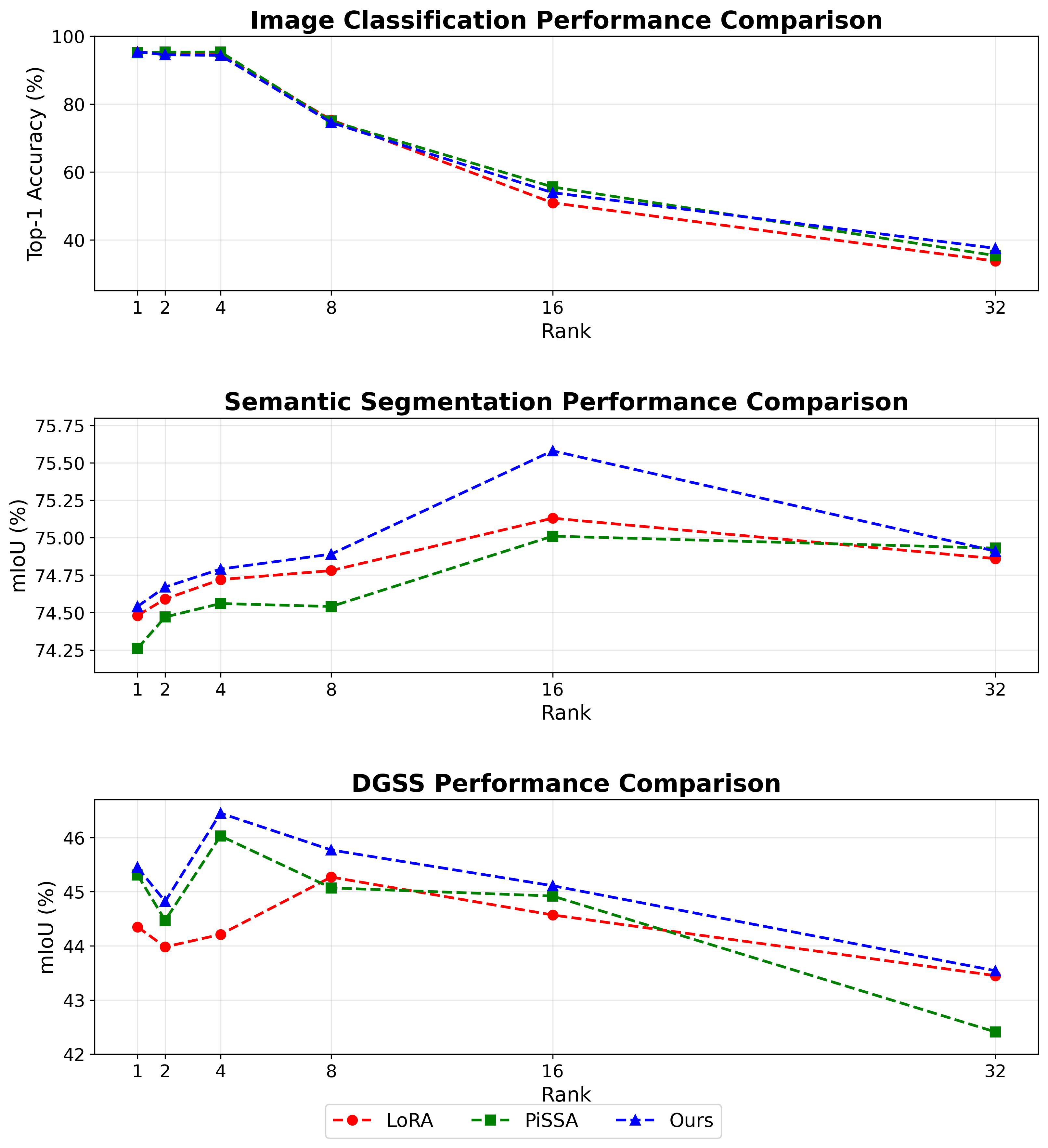}
    \caption{Comparison of different subspace tuning principles: LoRA performs fine-tuning by randomly assigning subspaces, PiSSA restricts subspaces to the principal components of weights. DROS achieves optimal matching.}
    \label{fig:rank}
\end{figure}

Fig. \ref{fig:rank} presents the performance trends of LoRA, PiSSA, and DROS under varying ranks across classification, segmentation, and DGSS tasks. DROS consistently achieves stronger performance under low-rank settings and exhibits a more favorable initialization behavior compared to baselines. Although performance differences diminish as rank increases, DROS maintains a consistent advantage in low-resource configurations, making it particularly suitable for computationally constrained remote sensing applications.
\subsubsection{Performance under limited labeled data}
To evaluate the effectiveness of DROS under data scarcity, all methods are fine-tuned with only 10\% of the training labels on DINOv2-L. As shown in Tab. \ref{tab:ablation_lowdata}, DROS attains the highest accuracy on all three benchmarks and the highest average metric of 67.29\%, ahead of both general-purpose PEFT methods such as LoRA, VPT, and PiSSA and remote sensing adapters such as Earth-Adapter. Notably, the margin of DROS over LoRA on RESISC45 grows from +0.04 under full supervision to +1.75 with only 10\% of the labels, and from +0.27 to +2.73 over PiSSA, indicating that the domain-aware subspace is particularly beneficial when supervision is scarce. The absolute margins over the best competing method remain modest, between 0.15 and 0.43 points, and label budgets below 1\% are not examined here; such extreme low-data settings are left to future work.
\begin{table}[ht]
\centering
\caption{Comparative results under 10\% training data on DINOv2-L.}
\label{tab:ablation_lowdata}
\resizebox{\columnwidth}{!}{
\begin{tabular}{lcccc}
\toprule
{\textbf{Method}} & {\textbf{RESISC45}} & {\textbf{Chesap.}} & {\textbf{U2R}} & {\textbf{Avg}} \\
\midrule
{FT} & 88.32 $\pm$ 0.71 & 69.72 $\pm$ 0.26 & 42.62 $\pm$ 0.59 & 66.89 \\
{FZ} & 70.00 $\pm$ 0.06 & 68.37 $\pm$ 0.32 & 42.23 $\pm$ 0.57 & 60.20 \\
{LoRA} & 87.19 $\pm$ 0.65 & 69.69 $\pm$ 0.79 & 42.60 $\pm$ 0.20 & 66.49 \\
{Earth-Adapter} & 88.68 $\pm$ 0.27 & 69.44 $\pm$ 0.63 & 42.20 $\pm$ 0.69 & 66.77 \\
{VPT} & 82.79 $\pm$ 2.08 & 68.45 $\pm$ 0.73 & 41.88 $\pm$ 0.14 & 64.37 \\
{PiSSA} & 86.21 $\pm$ 1.62 & 68.91 $\pm$ 1.13 & 42.49 $\pm$ 0.85 & 65.87 \\
{\textbf{DROS}} & \textbf{88.94 $\pm$ 0.10} & \textbf{70.15 $\pm$ 0.24} & \textbf{42.77 $\pm$ 0.40} & \textbf{67.29} \\
\bottomrule
\end{tabular}%
}
\end{table}
\subsubsection{Ablation of the relaxed orthogonal operator components}
The components of the relaxed orthogonal operator, parameterized by an orthogonal matrix $O$ together with two relaxation vectors $\alpha$ and $\beta$, are further ablated. As shown in Tab.~\ref{tab:ablation_operator}, omitting the operator yields an average metric of 71.50\%, while the $O$-only formulation obtains 71.71\%. The two single-sided variants obtain 71.44\% for $O$+$\alpha$ and 72.04\% for $O$+$\beta$, whereas the complete $O$+$\alpha$+$\beta$ formulation achieves the highest observed average of 72.24\%, including 75.58\% on Chesap.\ and 45.77\% on U2R. These results indicate a modest empirical advantage from retaining magnitude modulation on both sides of the orthogonal transformation. It is emphasized, however, that because the surrounding factors $A$ and $B$ are also trainable, scale-equivalent parameterizations exist and the dual-sided form does not enlarge the theoretical function class. It should therefore be interpreted as a structured optimization parameterization that explicitly separates pre- and post-reorientation magnitude modulation. The present ablation supports its downstream performance, but it does not establish universally faster convergence or provide a formal guarantee against scale drift; dedicated diagnostics of the individual scaling norms were not conducted.
\begin{table}[ht]
\centering
\caption{Ablation study of the relaxed orthogonal operator components on DINOv2-L.}
\label{tab:ablation_operator}
\resizebox{\columnwidth}{!}{
\begin{tabular}{lcccc}
\toprule
{\textbf{Operator}} & {\textbf{RESISC45}} & {\textbf{Chesap.}} & {\textbf{U2R}} & {\textbf{Avg}} \\
\midrule
{None (w/o operator)} & 95.18 $\pm$ 0.10 & 74.80 $\pm$ 0.19 & 44.52 $\pm$ 0.23 & 71.50 \\
{$O$} & 95.28 $\pm$ 0.10 & 75.07 $\pm$ 0.10 & 44.78 $\pm$ 0.61 & 71.71 \\
{$O$+$\alpha$} & 95.30 $\pm$ 0.05 & 74.91 $\pm$ 0.19 & 44.12 $\pm$ 0.98 & 71.44 \\
{$O$+$\beta$} & 95.33 $\pm$ 0.23 & 75.17 $\pm$ 0.11 & 45.61 $\pm$ 0.35 & 72.04 \\
{\textbf{$O$+$\alpha$+$\beta$}} & \textbf{95.37 $\pm$ 0.10} & \textbf{75.58 $\pm$ 0.13} & \textbf{45.77 $\pm$ 0.61} & \textbf{72.24} \\
\bottomrule
\end{tabular}%
}
\end{table}
\subsubsection{Ablation of the shared subspace in MM-DROS}
Which components of the relaxed orthogonal operator should be shared across modalities in MM-DROS is investigated next. As shown in Tab.~\ref{tab:ablation_mmshare}, the non-shared baseline achieves 65.41\% on BRIGHT, and sharing a subset of components yields gradual improvements. Sharing the full operator $O$+$\alpha$+$\beta$ yields the highest observed mean mIoU of 65.60\% among the tested configurations, suggesting a modest empirical benefit from sharing the complete transformation structure across modalities.
\begin{table}[ht]
\centering
\caption{Ablation study of the shared subspace in MM-DROS on BRIGHT.}
\label{tab:ablation_mmshare}
\begin{tabular}{lc}
\toprule
{\textbf{Shared Operator}} & {\textbf{BRIGHT}} \\
\midrule
{None (non-shared)} & 65.41 $\pm$ 0.84 \\
{$O$} & 65.46 $\pm$ 0.87 \\
{$O$+$\alpha$} & 65.47 $\pm$ 0.47 \\
{$O$+$\beta$} & 65.50 $\pm$ 0.71 \\
{\textbf{$O$+$\alpha$+$\beta$}} & \textbf{65.60 $\pm$ 0.84} \\
\bottomrule
\end{tabular}%

\end{table}
\subsection{Qualitative visualization on unseen wildfire scenes}
\begin{figure}[!tbp]
    \centering
    \includegraphics[width=\columnwidth]{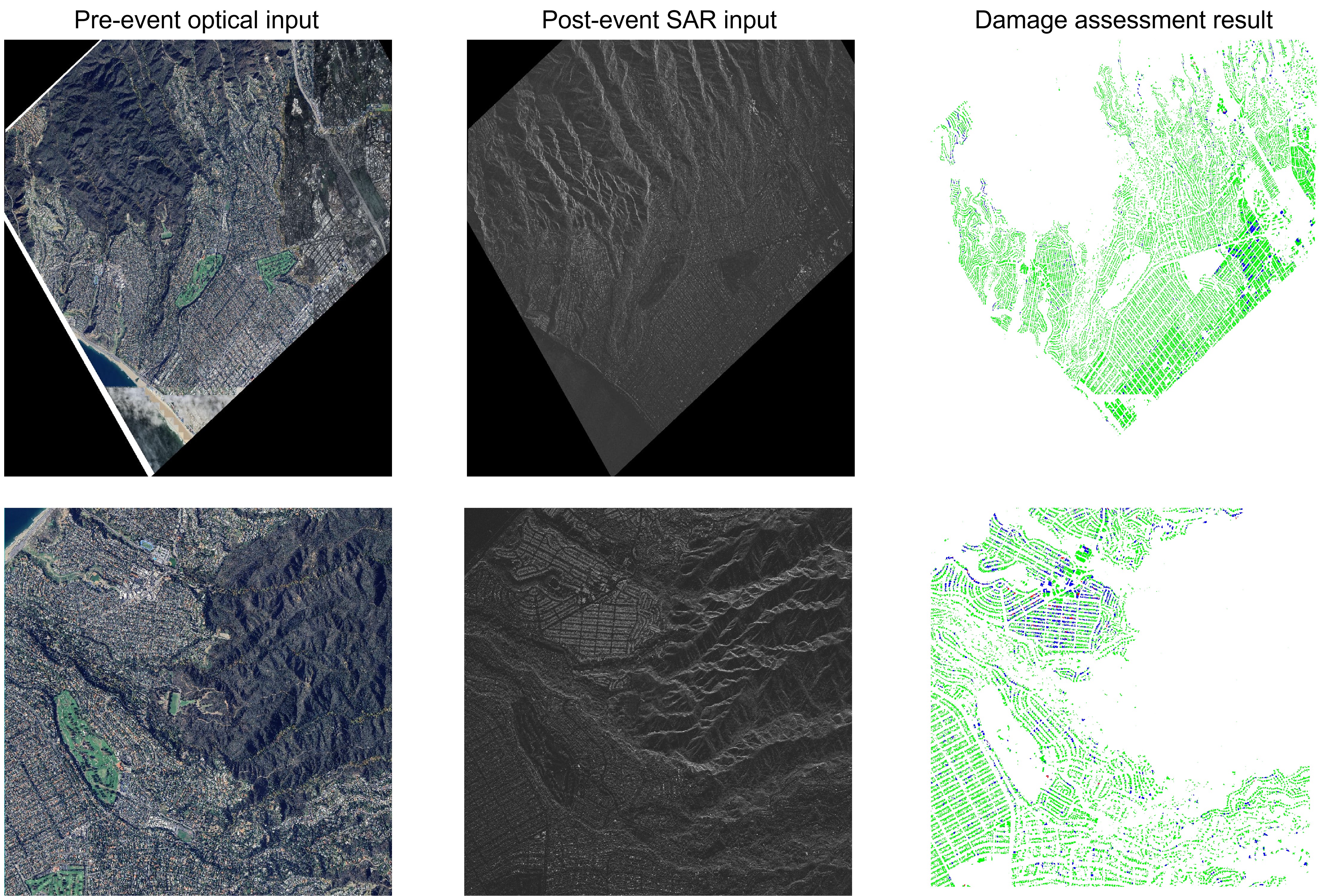}
    \caption{Qualitative case study of MM-DROS on the unseen Los Angeles wildfire scenario.}
    \label{fig:los}
\end{figure}

Fig. \ref{fig:los} provides a qualitative case study of MM-DROS on the Los Angeles wildfire scenario. This scenario involves substantial domain shifts, including complex pre-event urban vegetation and post-event SAR structural distortions caused by fire damage. Despite these challenges, MM-DROS produces coherent damage maps that distinguish intact and affected regions while reducing confusion caused by radar shadow effects, and temporal comparisons between January 8 and 9, 2025 illustrate the spatial localization of evolving damage regions. It is emphasized that this case study is presented solely to visualize the transfer behavior of the proposed framework on an unseen, highly non-stationary disaster scene; since no ground-truth damage annotation is available for this scenario, no quantitative metrics are reported and no comparison with other methods is made. It should therefore not be interpreted as quantitative evidence of zero-shot generalization.
\begin{figure*}[tb]
    \centering
    \includegraphics[width=\textwidth]{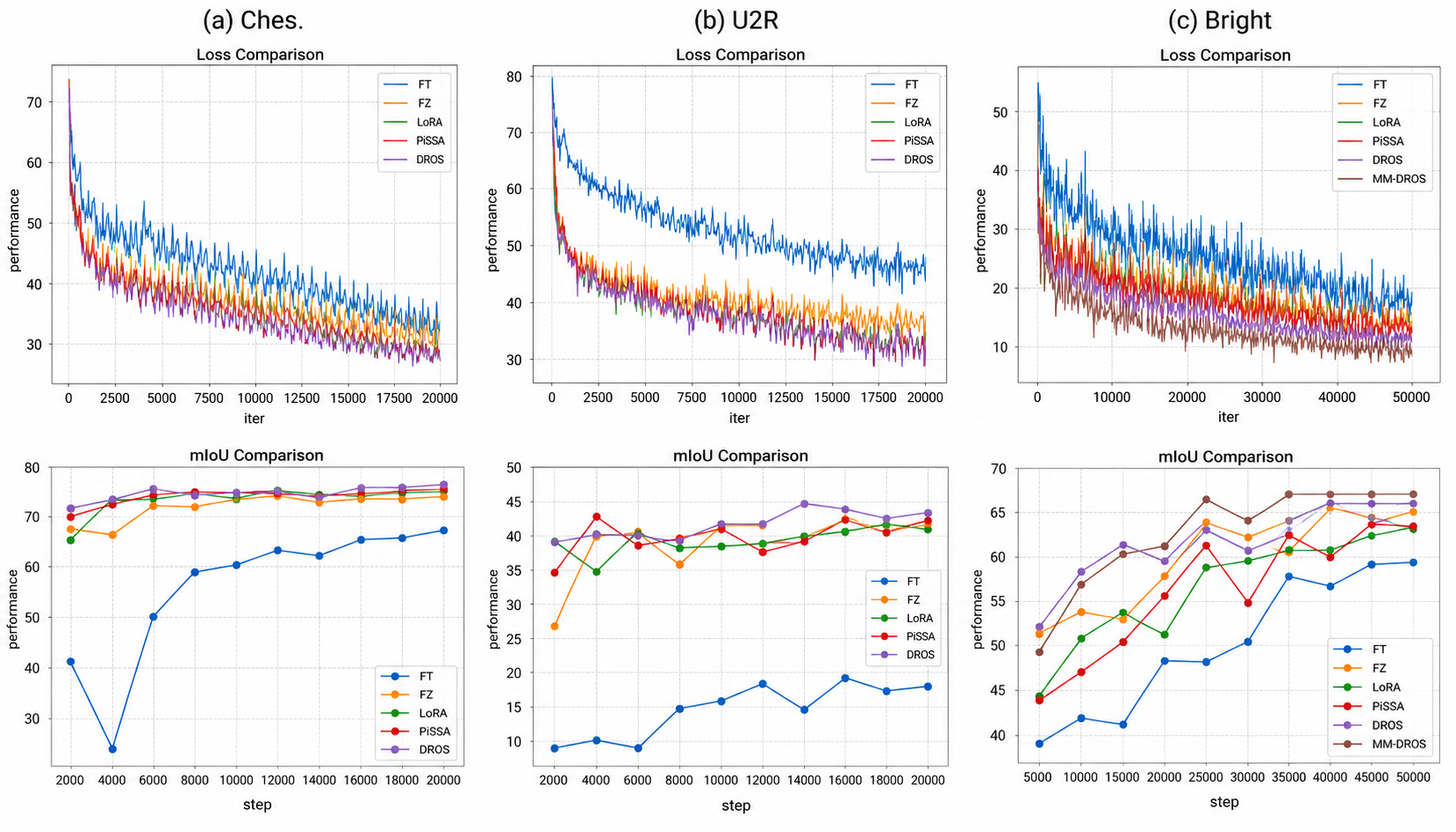}
    \caption{Comparison of loss curves and accuracy curves for different PEFT methods.}
    \label{fig:lossandmiou}
\end{figure*}
\subsection{Loss and convergence comparison}
Fig. \ref{fig:lossandmiou} illustrates the loss and mIoU trajectories across three benchmarks. Overall, DROS and MM-DROS reach a competitive performance level compared with LoRA and PiSSA, while consistently outperforming the full fine-tuning baseline. DROS does not universally achieve faster convergence: on in-domain segmentation with Chesapeake, the early-stage convergence of DROS is comparable to, or only marginally ahead of, LoRA. This is expected, because in-domain segmentation converges relatively easily and the strong decoder absorbs most of the task-specific variation, leaving little room for the adaptation subspace to shape the optimization trajectory; after sufficient training, LoRA and DROS settle into similar optima. The benefit of the domain-aware subspace is more evident where the domain shift is stronger. In the LoveDA DGSS setting, full fine-tuning shows decreasing loss but stagnant mIoU, whereas DROS maintains a consistent correlation between optimization and task performance. In the multimodal damage assessment task, MM-DROS demonstrates stable and gradual performance improvement over training iterations. Overall, DROS shows more favorable or stable optimization behavior in domain-shifted settings, rather than a consistent convergence advantage across all tasks.
\begin{table}[t]
\centering
\caption{Efficiency comparison on DINOv2-L across different methods. 
The number of learnable parameters (\#Params) and the peak GPU memory in GB are reported for each task.}
\label{tab:efficiency}
\resizebox{\columnwidth}{!}{
\begin{tabular}{c cc cc cc}
\toprule
\textbf{Method} 
& \multicolumn{2}{c}{\textbf{RESISC45}} 
& \multicolumn{2}{c}{\textbf{Chesap.}} 
& \multicolumn{2}{c}{\textbf{U2R}} \\
\cmidrule(lr){2-3} \cmidrule(lr){4-5} \cmidrule(lr){6-7}
& \textbf{\#Params} & \textbf{Mem. (GB)} 
& \textbf{\#Params} & \textbf{Mem. (GB)} 
& \textbf{\#Params} & \textbf{Mem. (GB)} \\
\midrule
FT    & 304.20M & 5.21 & 304.20M & 6.41 & 304.20M & 13.71 \\
FZ    & 0.00M   & 1.24 & 0.00M   & 2.42 & 0.00M   & 4.65 \\
LoRA  & 0.59M   & 2.75 & 6.21M   & 4.02 & 6.21M   & 10.77 \\
{Earth-Adapter} & 9.59M & 3.18 & 9.59M & 4.35 & 9.59M & 11.24 \\
{VPT}   & 0.49M & 2.68 & 6.29M & 3.95 & 1.57M & 10.58 \\
PiSSA & 0.59M   & 2.75 & 6.21M   & 4.02 & 6.21M   & 10.77 \\
DROS  & 0.59M   & 2.75 & 6.22M   & 4.03 & 6.22M   & 10.78 \\
\bottomrule
\end{tabular}%

}
\end{table}
\subsection{Efficiency Analysis}
The efficiency of different PEFT methods on the DINOv2-L backbone is further analyzed. As shown in Tab. \ref{tab:efficiency}, PEFT methods drastically reduce the number of trainable parameters compared to FT, demonstrating their effectiveness in parameter-efficient adaptation. Peak GPU memory consumption during training is also reported. All PEFT methods significantly reduce memory usage relative to FT, making them more suitable for resource-constrained scenarios. Across the corresponding DINOv2-L evaluations, DROS obtains the best average metric among the compared PEFT methods for classification, segmentation, and DGSS while maintaining a comparable parameter scale and peak training memory to lightweight approaches such as LoRA and PiSSA. The additional costs of DROS are confined to initialization and training: a one-time frozen-model forward pass over the representative training subset, covariance accumulation, a matrix inversion, and an SVD, together with the computation of the relaxed operator during training. After the operator is merged into the weight update, inference incurs no additional latency.
\subsection{Discussion}
\subsubsection{Task-dependent behavior and comparison}
The advantage of DROS is task-dependent. On the near-saturated classification benchmarks in Tab.~\ref{tab:cls_results}, most competitive methods already achieve 95--99\% accuracy, so the margins are small; on the MARS average, DROS (97.32\%) and Earth-Adapter (97.31\%) are nearly tied. DROS is the strongest parameter-efficient method for in-domain segmentation on both backbones, although full fine-tuning remains better on MARS. Its clearest advantage appears in DGSS (Tab.~\ref{tab:dg_results_final}), where it obtains the best average mIoU on both backbones and exceeds LoRA and PiSSA in every evaluated transfer setting. This pattern is consistent with the method design: covariance conditioning adapts the coordinate system to the observed training activations, while bottom-$r$ selection limits direct perturbation of dominant pretrained directions. Compared with LoRA~\cite{lora}, DROS explicitly conditions the low-rank subspace on downstream activations; unlike PiSSA~\cite{pissa}, it preserves rather than adapts the dominant components of the covariance-conditioned weight. OFT-type methods~\cite{oftv2} impose geometric transformations, whereas DROS applies a relaxed operator only within the selected data-conditioned subspace. Earth-Adapter~\cite{earth-adapter} remains competitive on individual datasets but retains additional modules at inference, while DROS is mergeable. DROS is therefore positioned as a domain-conditioned, mergeable adaptation method rather than a universally superior alternative: it requires one-time activation collection and SVD initialization but adds no inference latency after merging.

\subsubsection{Robustness analysis}
Robustness is evaluated only with respect to stochastic training variation and the domain shifts represented by DGSS. The main results report the mean $\pm$ standard deviation over three seeds; across the 20 dataset--backbone settings in Tabs.~\ref{tab:cls_results}--\ref{tab:dg_results_final}, DROS improves over PiSSA in 19 settings and over LoRA in 18, with two-sided sign-test values of $p\approx4\times10^{-5}$ and $p\approx4\times10^{-4}$, respectively. In DGSS, the subspace is constructed exclusively from the source-domain training split without labeled or unlabeled target samples. DROS achieves the best average mIoU on MARS (44.81\%) and DINOv2-L (55.32\%) and exceeds LoRA and PiSSA in every evaluated transfer setting. This evidence supports run-to-run stability and robustness to the evaluated cross-city and cross-scene shifts, but not to arbitrary sensors, adversarial perturbations, or synthetic corruptions.

\subsubsection{Limitations and scope}
DROS is not strictly plug-and-play because covariance accumulation, matrix inversion, and SVD add a one-time initialization cost, although a small representative subset is sufficient (Tab.~\ref{tab:ablation_subset}) and merging removes inference overhead. The gains remain task-dependent, and Fig.~\ref{fig:lossandmiou} does not support a universal convergence advantage. Moreover, the dual-sided scaling has scale-equivalent parameterizations because the surrounding low-rank factors are trainable; without diagnostics of the $\alpha$ and $\beta$ trajectories, no claim is made that it precludes scale drift. Label budgets below 1\%, broader multimodal settings, and controlled cross-temporal transfer remain future work.

\section{Conclusion}
This paper identifies subspace mismatch as a key bottleneck in PEFT under vast domain shifts. Existing PEFT methods, which rely on task-agnostic low-rank subspaces, are shown to be insufficient for capturing the distributional variations in remote sensing scenarios. To address this issue, DROS proposes to construct a data-conditioned adaptation subspace using second-order activation statistics derived from the downstream training data. A relaxed orthogonal formulation is further introduced to enable flexible transformations within the subspace, allowing improved adaptation while preserving pretrained knowledge. Building on this design, MM-DROS extends the framework to multimodal settings by sharing orthogonal basis transformations across modality-specific subspaces, enabling efficient cross-modal alignment. Extensive experiments across multiple remote sensing benchmarks demonstrate that the proposed framework achieves state-of-the-art performance without introducing additional inference cost. These results highlight the importance of explicitly modeling domain-aware subspaces for effective and efficient adaptation of foundation models under domain shifts. This work suggests that domain-aware subspace construction is a fundamental factor in bridging the gap between pretraining and downstream adaptation in remote sensing.

Admittedly, DROS is not without limitations. Estimating the activation covariance requires a one-time forward pass over a small representative subset, and subspace construction involves an SVD at initialization, so the method, like PiSSA, is not strictly plug-and-play; once the operator is merged, however, inference adds no overhead. Its gains are also modest on near-saturated benchmarks, where any fine-tuning strategy has little room to improve. These trade-offs, together with future directions such as cheaper covariance estimation, training with extremely scarce labels, and broader multimodal and cross-temporal settings, are discussed in Sec. IV-G. Learning the adaptation subspace from the data, rather than fixing it a priori, remains a promising direction for adapting foundation models to the geospatial domain.
\ifCLASSOPTIONcaptionsoff
  \newpage
\fi


\bibliographystyle{IEEEtran}
\bibliography{bibtex/bib/IEEEabrv,bibtex/bib/IEEEexample}


\begin{IEEEbiography}[{\includegraphics[width=1in,height=1.25in,clip,keepaspectratio]{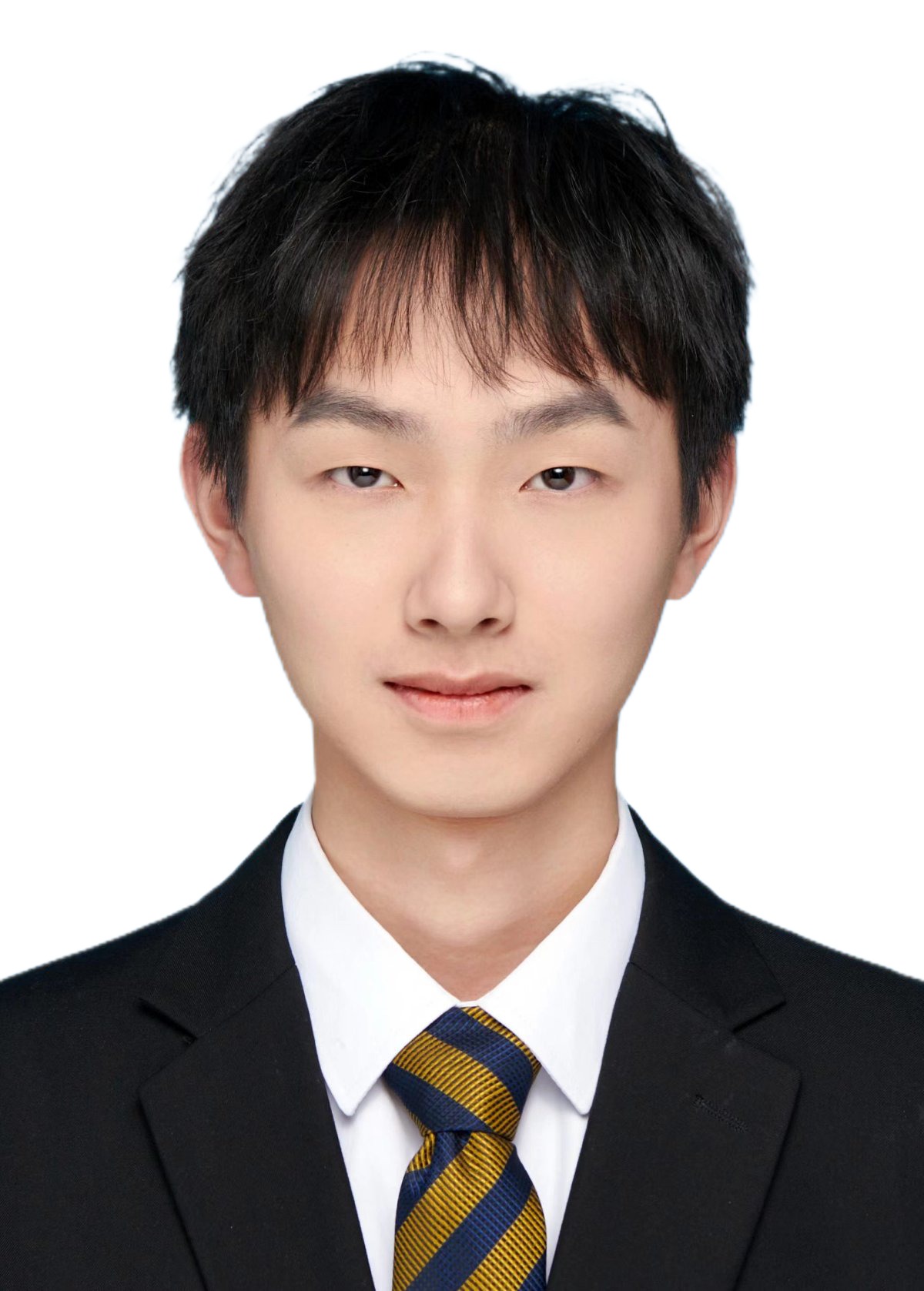}}]{Han Luo}
received the B.S. degree from the Division of Information Science and Engineering, Ocean University of China, Qingdao, China, in 2025. He is currently pursuing the M.S. degree with the State Key Laboratory of Information Engineering in Surveying, Mapping and Remote Sensing, Wuhan University, Wuhan, China. His major research interests include remote sensing image processing and computer vision, focusing on multimodal foundation models.
\end{IEEEbiography}
\begin{IEEEbiography}[{\includegraphics[width=1in,height=1.25in,clip,keepaspectratio]{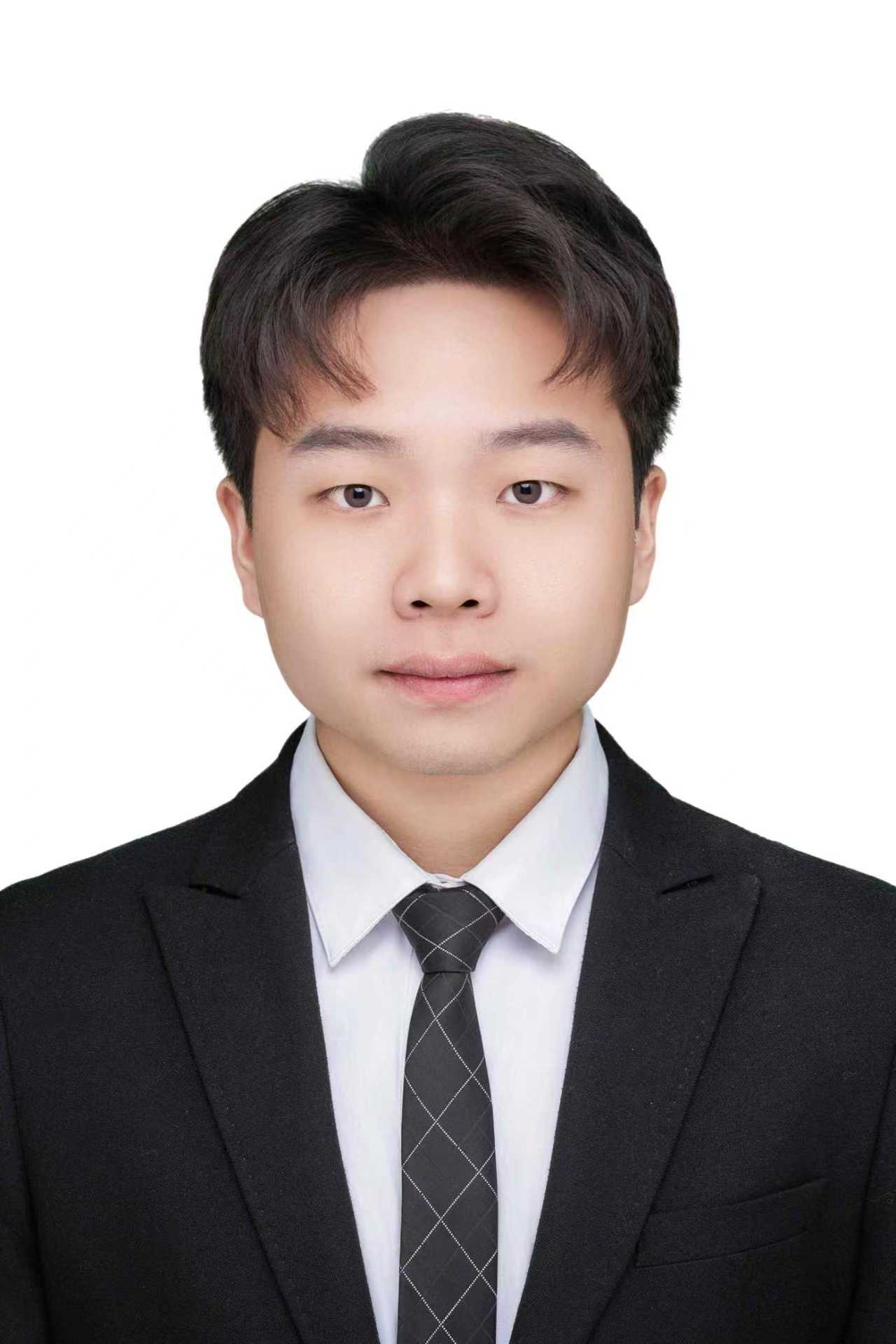}}]{Ruoyu Yang}
received the B.S. degree from the School of Resources and Civil Engineering, Northeastern University, Shenyang, China, in 2022. He is currently pursuing the M.S. degree with the State Key Laboratory of Information Engineering in Surveying, Mapping and Remote Sensing, Wuhan University, Wuhan, China. His major research interests include remote sensing image processing and computer vision, focusing on object vector extraction and road extraction.
\end{IEEEbiography}
\begin{IEEEbiography}[{\includegraphics[width=1in,height=1.25in,clip,keepaspectratio]{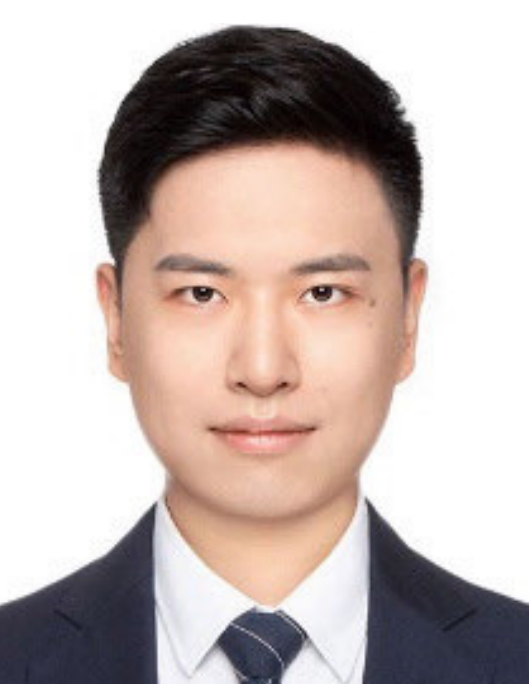}}]{Yinhe Liu}
 received the B.S. degree in resources and civil engineering from Northeastern University, Shenyang, China, in 2019, and the Ph.D. degree in photogrammetry and remote sensing from Wuhan University, Wuhan, China, in 2024. He is currently a postdoctoral researcher with the State Key Laboratory of Information Engineering in Surveying, Mapping and Remote Sensing (LIESMARS), Wuhan University, Wuhan, China.
His research interests focus on large-scale land-cover and land-use mapping and semantic change detection, with particular emphasis on developing advanced algorithms for geospatial intelligence applications.
\end{IEEEbiography}
\begin{IEEEbiography}[{\includegraphics[width=1in,height=1.25in,clip,keepaspectratio]{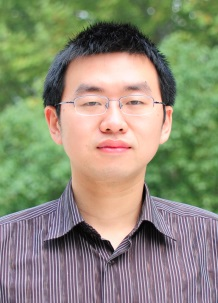}}]{Yanfei Zhong}
(Senior Member, IEEE) received the B.S. degree in information engineering and the Ph.D. degree in photogrammetry and remote sensing from Wuhan University, China, in 2002 and 2007, respectively.
Since 2010, He has been a Full professor with the State Key Laboratory of Information Engineering in Surveying, Mapping and Remote Sensing (LIESMARS), Wuhan University, China. He organized the Intelligent Data Extraction, Analysis and Applications of Remote Sensing (RSIDEA) research group. He has published more than 150 research papers in international journals, such as Remote Sensing of Environment, ISPRS Journal of Photogrammetry and Remote Sensing, IEEE Transactions on Geoscience and Remote Sensing. 
Dr. Zhong is a Fellow of the Institution of Engineering and Technology (IET). He is currently serving as an Associate Editor for the IEEE Journal of Selected Topics in Applied Earth Observations and Remote Sensing, and the International Journal of Remote Sensing.
\end{IEEEbiography}
\end{document}